\documentclass[11pt]{article}
\usepackage[top=2.54cm,left=2.54cm,right=2.54cm,bottom=2.54cm]{geometry}
\usepackage{amsmath,amsfonts}
\usepackage{algorithmic}
\usepackage{algorithm}
\usepackage{array}
\usepackage[caption=false,font=normalsize,labelfont=sf,textfont=sf]{subfig}
\usepackage{textcomp}
\usepackage{stfloats}
\usepackage{url}
\usepackage{verbatim}
\usepackage{graphicx}
\usepackage[square,numbers]{natbib}
\usepackage{tablefootnote}
\usepackage{xcolor}
\usepackage{hyperref}

\DeclareMathOperator*{\argmin}{arg\,min}
\usepackage{multirow}
\usepackage[normalem]{ulem}
\useunder{\uline}{\ul}{}
\usepackage{amsthm}

\newtheorem{assumption}{{Assumption}}
\newtheorem{theorem}{Theorem}
\newtheorem{lemma}{Lemma}

\newcommand{\be}{\begin{equation}}
\newcommand{\ee}{\end{equation}}

\providecommand{\keywords}[1]
{
  \small	
  \textbf{\textit{Keywords---}} #1
}

\begin{document}

\title{Smoothed Robust Phase Retrieval}

\author{Zhong Zheng and Lingzhou Xue \\
Department of Statistics, The Pennsylvania State University
}

\date{}

\maketitle
\renewcommand{\baselinestretch}{1.6}

\begin{abstract}
The phase retrieval problem in the presence of noise aims to recover the signal vector of interest from a set of quadratic measurements with infrequent but arbitrary corruptions, and it plays an important role in many scientific applications. However, the essential geometric structure of the nonconvex robust phase retrieval based on the $\ell_1$-loss is largely unknown to study spurious local solutions, even under the ideal noiseless setting, and its intrinsic nonsmooth nature also impacts the efficiency of optimization algorithms. This paper introduces the smoothed robust phase retrieval (SRPR) based on a family of convolution-type smoothed loss functions. Theoretically, we prove that the SRPR enjoys a benign geometric structure with high probability: (1) under the noiseless situation, the SRPR has no spurious local solutions, and the target signals are global solutions,
and (2) under the infrequent but arbitrary corruptions, we characterize the stationary points of the SRPR and prove its benign landscape, which is the first landscape analysis of phase retrieval with corruption in the literature. Moreover, we prove the local linear convergence rate of gradient descent for solving the SRPR under the noiseless situation. Experiments on both simulated datasets and image recovery are provided to demonstrate the numerical performance of the SRPR.
\end{abstract}


\section{Introduction}

Investigating the recovery of a signal vector of interest \(x \in \mathbb{R}^p\) from a set of sensing vectors \(a_1, \ldots, a_n \in \mathbb{R}^p\) and their linear transformations \(a_1^\top x, \ldots, a_n^\top x\) in the presence of noise is a critical research question in statistics and data science. There has been considerable interest in the estimation of this signal vector utilizing magnitude-based measurements \((a_1^\top x)^2, \ldots, (a_n^\top x)^2\) in the presence of potential errors and corruptions. This problem, commonly referred to as phase retrieval, plays a pivotal role in addressing the phase issue \citep{taylor2003phase, shechtman2015phase}. It finds applications across a wide spectrum of scientific applications, including X-ray crystallography \citep{miao1999extending}, optical physics \citep{millane1990phase}, coherent diffractive imaging using arrays and high-power sources \citep{chai2010array}, astronomy \citep{fienup1987phase}, and microscopy \citep{miao2008extending}.

Over the last decade, several pioneering papers have established the  theoretical foundation for the noiseless phase retrieval problem, which seeks to recover the signal vectors $x_\star$ or $-x_\star$ in $\mathbb{R}^p$ from a collection of magnitude-based measurements without any errors or corruptions:
\begin{equation}\label{intro_noiseless_phase}
b_{i}= (a_i^\top  x_\star)^2,
\end{equation}
for each $i \in [n] := \{1, 2, \ldots, n\}$. Here, $b_1, \ldots, b_n$ represent non-negative measurements. It is important to note that \eqref{intro_noiseless_phase} is a nonconvex problem recognized as being NP-hard \citep{fickus2014phase}. To address this challenge, \cite{candes2013phaselift} introduced the PhaseLift method, employing a convex relaxation strategy based on semidefinite programming (SDP) to solve a trace-norm minimization problem. However, as the dimension of the signal vector increases, SDP-based methods tend to become computationally infeasible \citep{candes2015phase}.
In response to the limitations of SDP-based methods, the literature has seen alternative approaches based on the non-convex optimization, including the Wirtinger flow \citep{candes2015phase}, truncated Wirtinger flow \citep{chen2017solving}, thresholded Wirtinger flow \citep{cai2016optimal}, truncated amplitude flow \citep{wang2017solving}, and others. These non-convex methods have utilized spectral initialization \citep{candes2015phase} to facilitate global convergence. 
Recently, the seminal paper \cite{sun2018geometric} demonstrated that the least-squares formulation for recovering the signal from \eqref{intro_noiseless_phase} exhibits a benign global geometry (also known as function landscape) ensuring that, with high probability, the non-convex problem has no spurious local minimizers and can be globally optimized through efficient iterative methods, even with random initialization. Further, \cite{chen2019gradient} showed that vanilla gradient descent is effective and provided a comprehensive analysis of the optimization trajectory, and \cite{cai2022solving} illustrated that the smoothed amplitude flow model shares similar properties.

In real-world applications, it is crucial not to overlook the influence of noise on phase retrieval. Especially, we should consider the presence of additive errors and infrequent corruptions affecting magnitude-based measurements, which may arise due to equipment limitations, measurement errors, extreme occurrences, and so on. The noisy magnitude-based measurements can be formalized as follows: for each $i \in [n]$,
\begin{equation}\label{intro_corrupted_phase}
b_i = \begin{cases}
  (a_i^\top x_\star)^2 + \tau_i, & \text{if the $i$-th measurement is uncorrupted}, \\
  \xi_i, & \text{otherwise}.
\end{cases}
\end{equation}
Here, $\tau_i$ represents the additive errors bounded in $\mathbb{R}$, and $\xi_i$ signifies the infrequent yet arbitrary corruptions in $\mathbb{R}_+$. The set $\mathbb{I}_O = \{i \in [n]: \text{the $i$-th measurement is corrupted}\}$ denotes the indices of outliers, and $\mathrm{card}(\mathbb{I}_O)$ is the cardinality of $\mathbb{I}_O$. We define $p_{\text{fail}} = \mathrm{card}(\mathbb{I}_O) / n$ as the proportion of corrupted measurements. It is essential to note that $p_{\text{fail}}$ remains a relatively small quantity due to the infrequency of corruption. However, corruptions $\xi_i$'s can be unbounded and may behave adversarially. Furthermore, independence between $\{\xi_i\}_{i\in \mathbb{I}_O}$ and $\{a_i\}_{i=1}^n$ may not hold, introducing additional challenges to signal recovery.

Model \eqref{intro_corrupted_phase} provides a general framework for signal recovery from noisy magnitude-based measurements. This framework includes various specialized scenarios as special examples. Specifically, the noiseless phase retrieval problem, as formulated in \eqref{intro_noiseless_phase}, can be perceived as a particular instance of \eqref{intro_corrupted_phase}, characterized by $p_{\text{fail}} = 0$ and $\tau_i = 0$ for all $i \in [n]$. Moreover, phase retrieval in the presence of bounded noise, as investigated by \cite{zhang2017fast}, emerges as another special case of \eqref{intro_corrupted_phase} when $p_{\text{fail}} = 0$, indicating the absence of outliers. Conversely, the robust phase retrieval problem \citep{duchi2019solving} aligns with \eqref{intro_corrupted_phase} under the situation that $\tau_i = 0$ for all $i \in [n]$, focusing on the robustness against corruptions. Additionally, the approach for non-convex phase retrieval with outliers, as proposed by \cite{zhang2016provable}, resonates with \eqref{intro_corrupted_phase} and accommodates both dense bounded noise and sparse arbitrary outliers: $b_i = (a_i^\top  x_\star)^2 + \tau_i + \xi_i$ for each $i \in [n]$.

To solve the problem of phase retrieval with corruptions in Model~\eqref{intro_corrupted_phase},
the objective remains to recover the signal vectors $x_\star$ or $-x_\star$ using the observations $\{(a_i, b_i)\}_{i=1}^n$. In this vein, \cite{zhang2016provable} extended the methodologies of \cite{chen2017solving} to propose the median truncated Wirtinger flow. \cite{duchi2019solving} explored robust phase retrieval employing the $\ell_1$-loss, formally defined as
\begin{equation}
    \label{duchi_l1_ori}
    \min_{x \in \mathbb{R}^p} F(x) := \min_{x \in \mathbb{R}^p}\frac{1}{n} \sum_{i=1}^{n} \left| (a_i^\top  x)^2 - b_i \right|.
\end{equation}
\cite{duchi2019solving} introduced the proximal linear (PL) algorithm to tackle this nonconvex optimization challenge, demonstrating that the $\ell_1$-loss enhances success rates in comparison to the median truncated Wirtinger flow  \cite{zhang2016provable}. Furthermore, they established the properties of sharpness and weak convexity for $F(x)$, alongside the local convergence of the PL algorithm. Subsequent research, targeting the optimization problem \eqref{duchi_l1_ori}, has expanded upon these findings. \cite{davis2018subgradient, davis2020nonsmooth} investigated the local convergence of subgradient algorithms. Recently, \cite{zheng2023new} proposed the inexact proximal linear (IPL) algorithm and studied its local convergence.

Despite these advancements, the essential geometric structure of the robust phase retrieval is largely unknown to study spurious local solutions. This contrasts with the works of \cite{sun2018geometric, chen2019gradient, cai2022solving}, which provided insights into the landscape of smooth objective functions in phase retrieval under noiseless conditions. Notably, even under the ideal noiseless setting, \cite{duchi2019solving} did not conduct landscape analysis to understand the stationary points of the nonconvex and nonsmooth optimization. \cite{davis2020nonsmooth} only defined and analyzed stationary points in the context of subgradient, and whether they are local minimums is still unknown. Moreover, under the situation with corruption, we cannot find any existing work to study the landscape of any objective function in the literature. As a result, there is no theoretical guarantee for robust phase retrieval to avoid spurious local minimums using random initialization. 

Another concern arises from the intrinsic nonsmooth nature of  \eqref{duchi_l1_ori} that significantly impacts the efficiency of optimization algorithms including the PL method \cite{duchi2019solving} and the IPL method \cite{zheng2023new}. Both approaches necessitate iteratively solving nonsmooth subproblems using the proximal operator graph splitting (POGS) algorithm \cite{parikh2014block} or the accelerated proximal gradient algorithm, which invariably leads to diminished optimization efficiency. Moreover, as pointed out by \cite{davis2018subgradient, davis2020nonsmooth}, the subgradient algorithms for solving \eqref{duchi_l1_ori} are sensitive to the step size selection, and there are no practically reliable methods for choosing the step size.

All these challenges stem from the non-differentiability of the $\ell_1$-loss employed in \eqref{duchi_l1_ori}, which limits the technical tools for analyzing function landscapes and complicates the optimization process. In our paper, we propose the smoothed robust phase retrieval (SRPR) method by minimizing the smoothed surrogate $F_\delta(x)$ based on the convolution-type smoothing, which involves the convolution of the $\ell_1$-loss with a kernel function $K_\delta(\cdot)$ and its bandwidth $\delta>0$. This smoothing idea was first introduced by \cite{fernandes2021smoothing} and then investigated by \cite{he2021smoothed}, \cite{tan2021high}, \cite{jiang2021smoothing}, and \cite{man2022unified} to overcome the lack of smoothness in quantile regression and improve optimization efficiency while achieving the desired convergence rate and oracle property.

We demonstrate that, with high probability, $F_\delta(x)$ not only satisfies the weak convexity property similar to $F(x)$ but also satisfies generalized sharpness in the noiseless scenario. Furthermore, this approach facilitates the utilization of gradient and Hessian information, providing a convenient mathematical structure. These properties are crucial for determining the convergence rates of gradient-based optimization algorithms and for conducting landscape analyses. By directly minimizing $F_\delta(x)$ in the absence of noise, exact signal recovery is achievable. Additionally, our methodology extends to scenarios involving corrupted measurements, illustrating its versatility and applicability. The SRPR approach has at least the following contributions to the field, enhancing both the theoretical and practical aspects of phase retrieval when compared to existing methodologies.

Firstly, the smoothness of our approach enables the application of gradient-based algorithms to minimize the nonconvex function $F_\delta(x)$. We demonstrate that under the noiseless situation, the minimization of $F_\delta(x)$ through general first-order methods allows SRPR to achieve local linear convergence. This marks a significant improvement over the sublinear convergence rates observed with the PL \citep{duchi2019solving} and IPL \citep{zheng2023new} algorithms. The attainment of a linear rate is facilitated by the Polyak-Lojasiewicz condition \citep{karimi2016linear} that holds locally, circumventing the need for local convexity. Additionally, our method benefits from the potential for acceleration and line search strategies, offering clear advantages over subgradient algorithms \citep{davis2018subgradient,davis2020nonsmooth}, which encounter challenges in step size tuning. In numerical experiments, SRPR demonstrates superior efficiency relative to competing algorithms.

Secondly, in the context of noiseless phase retrieval, we conduct landscape analysis to show that $F_\delta(x)$, with high probability, exhibits a benign landscape and has no spurious local minima. This analysis underscores the efficacy of random initialization in achieving exact recovery through the minimization of $F_\delta(x)$. To the best of our knowledge, this is the first instance of a benign landscape being identified for a robust objective function in the domain of noiseless phase retrieval, as previously discussed works \citep{duchi2019solving} and \cite{davis2020nonsmooth} have only considered the local behaviors of optimization algorithms and the locations of stationary points.

Thirdly, in the presence of corrupted measurements, we prove that when $p_{\text{fail}}$ is small and bounded noises are absent, with high probability, the landscape of $F_\delta(x)$ remains benign outside a vicinity around $\{x_\star, -x_\star\}$ with a radius proportional to $\delta p_{\text{fail}}/(1-p_{\text{fail}})$. Similar results hold for the general Model \eqref{intro_corrupted_phase}. 
To the best of our knowledge, this represents the first result of benign function landscapes in the context of phase retrieval with corruptions. 

The rest of this paper is organized as follows. Section \ref{sec:methodology} introduces the methodology of SRPR. Section \ref{sec:landscape_noiseless} focuses on the noiseless landscape analysis. Section \ref{sec:landscape_noise} focuses on the empirical landscapes with corruption. Section \ref{sec:numerical} presents the numerical experiments. Section \ref{conclusion} includes a few concluding remarks. The complete proofs are presented in the supplement.

\section{Methodology}\label{sec:methodology}

This section first presents the general framework for smoothed robust phase retrieval in Subsection \ref{subsec:methodology:gen_frame} and then discusses the convolution-type smoothed loss functions in Subsection \ref{subsec:methodology:conv_smoothing}. Subsection \ref{subsec:methodology:properties_obj} presents the properties of the smoothed objective function. Subsection \ref{subsec:methodology:rate} discusses the optimization algorithms.

Before proceeding, we introduce some useful notations. Denote $\mathbf{1}[\cdot]$ as the indicator function. Let $B_p(x,M) = \{y\in\mathbb{R}^p: \|y-x\|_2\leq M\}$. Define $\mathbb{E}_s A = \frac{1}{s}\sum_{i = 1}^s A_i$, and $I_p$ denotes a $p\times p$ identity matrix. Let $e_i\in\mathbb{R}^p$ represent a vector whose $i-$th element is 1 with all other elements taking the value 0. In addition, throughout the paper, we use $c,c',C,C',...$ to denote generic positive constants, though the actual values may vary on different occasions.

\subsection{A General Framework}\label{subsec:methodology:gen_frame}

In the robust phase retrieval problem \eqref{intro_corrupted_phase}, we observe vectors $\{a_i\}$ and non-negative scalars $\{b_i\}$ and aim to recover the true phases $\{x_\star, -x_\star\}$\footnote{Due to the quadratic relationship in \eqref{intro_corrupted_phase}, both $x_\star$ and $-x_\star$ are true signal vectors.}. %The formal statistical framework is articulated as follows:

\begin{assumption}\label{ass_data}
Let $a_{1}, a_{2}, \ldots, a_{n_1+n_2}$ be random vectors in $\mathbb{R}^p$ that are independently and identically distributed with the mean zero and covariance matrix $\Sigma_p$, satisfying $u^\top\Sigma_pu \leq \lambda_M\|u\|_2^2$ for all $u \in \mathbb{R}^p$, where $\lambda_M$ is a positive constant. Here, $n_1, n_2$ are non-negative integers, and $p \geq 3$ is a positive integer. The true phase $x_\star \in \mathbb{R}^p \setminus \{0\}$ and
\begin{equation}
    b_i = [(a_i^\top x_\star)^2 + \tau_i]\mathbf{1}[i \in [n_1]] + \xi_i\mathbf{1}[i \in [n_1+n_2] - [n_1]],\ \forall\ i \in [n_1+n_2].
\end{equation}
The $\tau_i$'s are random variables in $\mathbb{R}$ with $|\tau_i| \leq \gamma$ for any $i \in [n_1]$. The $\xi_i$'s are non-negative random variables without further assumptions. 
\end{assumption}

In Assumption \ref{ass_data}, $n = n_1 + n_2$ represents the total number of measurements, and $p_{\text{fail}} = \frac{n_2}{n_1 + n_2}$ denotes the proportion of corrupted measurements. The sets $\mathbb{I}_1 = [n_1]$ and $\mathbb{I}_2 = [n] - [n_1]$ correspond to the inliers and outliers among the measurements, respectively. In the special case where $\gamma = p_{\text{fail}} = 0$, the model reverts to the noiseless phase retrieval scenario as depicted in \eqref{intro_noiseless_phase}. Our goal is to recover the true signal vectors $x_\star$ or $-x_\star$ from the observed $\{(a_i, b_i)\}_{i=1}^n$, without prior knowledge of which measurements are corrupted. This setup ensures the mutual independence of $\{a_i, i \in [n_1]\}$ and $\{a_i, i \in [n_1 + n_2] - [n_1]\}$. We will introduce stronger assumptions on the distribution of $a_i$ later. The bounded noise variables $\tau_i$s require no additional distributional assumptions, whereas no assumptions are made regarding the distribution of $\{\xi_i\}_{i=1}^n$ or its independence with respect to $\{a_i\}$, allowing for potentially heavy-tailed and adversarially distributed corruption. This assumption is  more general then the model assumption $M_2$ of \cite{duchi2019solving}, which requires independence between $\{\xi_i, i \in [n_1 + n_2] - [n_1]\}$ and $\{a_i, i \in [n_1]\}$. For simplicity, in subsequent discussions, we will refer to a random vector $a$ to denote the generic distribution of the $a_i$s.

As shown in \cite{duchi2019solving}, when $\gamma = 0$, exact recovery is achievable in the optimization problem \eqref{duchi_l1_ori} employing an $\ell_1$-loss, indicating that $\{x_\star, -x_\star\} = \argmin_{x \in \mathbb{R}^p} F(x)$, with high probability, under certain assumptions. To refine this approach, we employ convolution-type smoothing, focusing on the problem formulated as:
\begin{equation}\label{loss_smoothed}
\min_{x \in \mathbb{R}^p} F_\delta(x) = \frac{1}{n}\sum_{i=1}^{n}l_\delta\left((a_i^\top x)^2 - b_i\right),
\end{equation}
where
\begin{equation}\label{conv_typed_smoothed_loss}
    l_\delta(x) = (l * K_\delta)(x) = \int_{-\infty}^{+\infty} |y| K_\delta(x - y) dy,
\end{equation}
$K(x)$ represents a kernel function, $\delta > 0$ is the bandwidth, $*$ denotes the convolution operation, and $K_\delta(x) = \frac{1}{\delta}K(\frac{x}{\delta})$. In this context, $l_\delta(x)$ serves as a smoothed loss function, substituting $l(x)$ for $F(x)$. The adoption of convolution-type smoothing is motivated by gradient approximations: $\nabla F_\delta(x)$ closely mirrors the subgradient of $F(x)$ because $\lim_{\delta \rightarrow 0} l'_\delta(x) = \text{sign}(x) \in \partial |x|$, thereby suggesting that the statistical behavior of $F_\delta(x)$ will approximate that of $F(x)$. Moreover, the smoothness of $F_\delta(x)$ significantly benefits gradient-based optimization and landscape analysis. Provided some mild conditions on the kernel function $K(x)$, both $l'_\delta(x)$ and $l''_\delta(x)$ are well-defined, ensuring that $F_\delta(x)$ is smooth and equipped with gradients and Hessians for both optimization and landscape analysis purposes.

\subsection{Convolution-type Smoothing}\label{subsec:methodology:conv_smoothing}
In this subsection, we elaborate on the convolution-type smoothed loss function as defined in \eqref{conv_typed_smoothed_loss}.  Additionally, we introduce the notations $\Tilde{K}(x) = \int_{-\infty}^x K(y) dy$ and $\Tilde{K}_\delta(x) = \Tilde{K}(x/\delta)$ for the integrated kernel function and its scaled version, respectively. To facilitate our analysis, we adhere to the subsequent assumption regarding the kernel function $K(x)$ throughout this paper:

\begin{assumption}\label{ass_kernel}
The mapping $K:\mathbb{R}\rightarrow \mathbb{R}$ satisfies the following conditions:
\begin{enumerate}
    \item [(a)] $K(\cdot)$ is a kernel, namely, $K(x)\geq 0,K(x)=K(-x),\forall\ x\in \mathbb{R}$ and $\int_{-\infty}^\infty K(x)dx = 1$.
    \item [(b)] $K(x)$ is bounded, Lipschitz continuous and $K(0)>0$.
    \item [(c)] Let $\bar{K}(x) = xK(x)$. $\bar{K}(x)$ is bounded, Lipschitz continuous and  $\int_{-\infty}^{+\infty}\left|\bar{K}(x)\right|dx<\infty$.
\end{enumerate}
\end{assumption}
Next, we give some examples of the kernel $K(x)$ that satisfies Assumption \ref{ass_kernel}.
\begin{itemize}
\item Gaussian kernel $K(x)=(2\pi)^{-1/2}e^{-x^{2}/2},$ which generates $l_{\delta}(x)=\delta l^{\mathrm{G}}(x/\delta)$ with $l^{\mathrm{G}}(x):=(2/\pi)^{1/2}e^{-x^{2}/2}+x\{1-$ $2\Phi(-x)\}$ and $\Phi(\cdot)$ being the CDF of $N(0,1)$.
\item Logistic kernel $K(x)=e^{-x}/\left(1+e^{-x}\right)^{2},$ which generates $l_{\delta}(x)=\delta l^{\mathrm{L}}(x/\delta)$ with $l^{\mathrm{L}}(x)=x+2\log\left(1+e^{-x}\right)$.
\item Epanechnikov kernel $K(x)=(3/4)\left(1-x^{2}\right) \mathbf{1}(|x| \leq 1),$ which generates $ l_{\delta}(x)=\delta l^{\mathrm{E}}(x/\delta)$ with $l^{\mathrm{E}}(x)=\left(3x^{2}/4-x^{4}/8+3/8\right) \mathbf{1}(|x| \leq 1)+|x| \mathbf{1}(|x|>1)$.
\item Triangular kernel $K(x)=(1-|x|) \mathbf{1}(|x\leq 1)$, which generates $l_{\delta}(x)=\delta l^{\mathrm{T}}(x/\delta)$ with $l^{\mathrm{T}}(u)=(x^{2}-|x|^{3}/3+1/3)\mathbf{1}(|x| \leq 1)+|x| \mathbf{1}(|x|>1)$.
\item $K(x)=\frac{1}{2(x^2+1)^{3/2}},$ which generates the Pseudo-Huber loss $l_{\delta}(x) = \sqrt{x^2+\delta^2}$.
\end{itemize}

In practical implementations, $\delta$ ought to be a small constant so that $l_\delta(x)$ approximates the $\ell_1$-loss, following the ideas introduced in smoothed quantile regression \citep{he2021smoothed}.
Following Assumption \ref{ass_kernel}, we can derive several properties of $l_\delta(x)$, which are briefly discussed here for space consideration. {More details are summarized in Lemma \ref{lemma_property_huber}
of Appendix \ref{appendix:property_hubers} in the supplement.
} Similar to \cite{he2021smoothed}, we prove that $l_\delta(x)$ is convex and has continuous first-order and second-order derivatives, namely, $
l'_\delta(x) = 2\Tilde{K}_\delta(x) - 1$ and $l''_\delta(x) = 2K_\delta(x).$ This ensures the existence of the gradient and Hessian for $F_\delta(x)$. Moreover, note that
$
l_\delta(x) - l_\delta(0) \geq \min\left\{\frac{C_1}{\delta}x^2, C_2|x|\right\}, \forall x \in \mathbb{R}^p,
$ for some positive constants $C_1$ and $C_2$. This relation suggests that $l_\delta(x)$ exhibits characteristics akin to the $\ell_2$-loss for values of $x$ near $0$, and resembles the $\ell_1$-loss for values of $x$ distant from $0$. This bound acts as the foundation for the generalized sharpness condition to $F_\delta(x)$ in the absence of noise, a topic to be elaborated upon in the subsequent subsection.

\subsection{Properties of the Smoothed Objective Function}\label{subsec:methodology:properties_obj}
In this subsection, we discuss the properties of $F_\delta(x)$, which will indicate the similarities between $F(x)$ and $F_\delta(x)$ and how minimizing $F_\delta(x)$ works for robust phase retrieval. We first introduce two parallel assumptions given in \cite{duchi2019solving}. Assumption \ref{ass_direct} claims that the distribution of $a_i$ is non-degenerate in all the directions, and Assumption \ref{ass_subg} describes the tail-behavior of the distribution of $a_i$.

\begin{assumption}
\label{ass_direct}
There exists two positive constants $\kappa_{st}$ and $p_{st}$ such that, for any $i\in [n]$ and $u,v\in \mathbb{R}^p$ such that $\|u\|_2 = \|v\|_2 =1$, we have
$P\left(\min\{|a_i^\top u|,|a_i^\top v|\}\geq \kappa_{st}\right)\geq p_{st}.$
\end{assumption}

\begin{assumption}
\label{ass_subg}
The vectors $a_i$'s are sub-Gaussian with parameter $c_s\geq 1$, which means that
$$\inf \left\{t>0: \mathbb{E} \exp \left((a_i^\top u)^{2} / t^{2}\right) \leq 2\right\}\leq c_s\sqrt{u^\top\Sigma_p u},\forall i\in [n],u\in \mathbb{R}^p.$$
\end{assumption}

With the same Assumptions, we can prove two properties for $F_\delta(x)$ which are comparable to properties for $F(x)$ as follows. The first one will be generalized sharpness.
\begin{lemma}\label{lemma_sharpness_smooth}
 (Generalized Sharpness) Under Assumptions \ref{ass_data}, \ref{ass_kernel} and \ref{ass_direct}, when $p_{\text{fail}} = \gamma = 0$ and $n\geq c p$, with probability at least $1-C\exp(-c'n)$, we have
 \begin{equation}\label{ineq_gen_sharp}
     F_\delta(x)-F_\delta(x_\star)\geq \lambda_s\min\{\Delta(x),\frac{1}{\delta}\Delta^2(x)\},\forall x\in \mathbb{R}^p.
 \end{equation}
where $\Delta(x) = \min\{\|x-x_\star\|_2,\|x+x_\star\|_2\}$, and $c,c',C,\lambda_s$ are constants independent of $\delta$.

\end{lemma}

In the literature, the sharpness of a function $f(x)$ bounds the difference between $f(x)$ and its global minimum by the distance between $x$ and its nearest global minimum point. This concept is prevalent in the optimization problems characterized by equation-solving aspects and nonsmooth objective functions. For more details, please see Section 1 of \cite{Roulet2017sharpness}. The generalized sharpness condition extends the concept of sharpness for $F(x)$, as outlined in Corollary 3.1 of \cite{duchi2019solving}, which is the special example of \eqref{ineq_gen_sharp} when $\Delta(x)\ge \delta$:
\begin{equation}\label{sharpness_F_pre}
    F(x) - F(x_\star) \geq \lambda_s \Delta(x), \forall x \in \mathbb{R}^p.
\end{equation} 
\cite{duchi2019solving} and \cite{Mendelson2014sharpness} proved that \eqref{sharpness_F_pre} holds with high probability for the noiseless robust phase retrieval. \cite{duchi2019solving}, \cite{davis2018subgradient}, \cite{davis2020nonsmooth}, and \cite{zheng2023new} leveraged \eqref{sharpness_F_pre} to show the convergence rate for their algorithms. 

The second one is the weak convexity that relies on the Assumption \ref{ass_subg}.
\begin{lemma}\label{lemma_smooth_weak_convexity}
(Weak Convexity) Given Assumptions \ref{ass_data}, \ref{ass_kernel}, and \ref{ass_subg}, and assuming $n \geq cp$, the inequality below is satisfied with a probability of at least $1-C\exp(-c'n)$:
\begin{equation}\label{rel_smooth_weak}
    F_\delta(y) - F_\delta(x) \geq \nabla F_\delta(x)^\top (y-x) - \frac{\rho}{2}\|y-x\|_2^2, \quad \forall x, y \in \mathbb{R}^p.
\end{equation}
Here, $\rho, C, c, c'$ are positive constants independent of $\delta$, $K(x)$, or the value of $p_{\text{fail}}$.
\end{lemma}
This relationship also holds when substituting $F_\delta(x)$ with $F(x)$ and replacing the gradient with the subgradient. Weak convexity, a notion extensively recognized in optimization literature, serves as a generalization of convexity. It necessitates that $f(x) + \frac{\rho}{2}\|x\|_2^2$ be convex for a certain positive constant $\rho$. Various definitions for weak convexity are discussed throughout the literature. The formulation presented in Lemma \ref{lemma_smooth_weak_convexity} aligns with that in \cite{davis2020nonsmooth}. Furthermore, Corollary 3.2 in \cite{duchi2019solving} defines weak convexity by bounding the deviation between $F(x)$ and its linear approximation, requiring the existence of a positive constant $\rho > 0$ such that:
$$\left|F(x) - \frac{1}{n}\sum_{i=1}^n \left|(a_i^\top y)^2-b_i+2(a_i^\top y)a_i^\top (x-y)\right|\right| \leq \frac{\rho}{2}\|y-x\|_2^2, \quad \forall x, y \in \mathbb{R}^p.$$

Next, under the case $\gamma = 0$, we discuss how generalized sharpness and weak convexity lead to the robustness of using $F_\delta(x)$. As $l_\delta(x)$ is an approximation of $l(x)$, we can expect the approximation in function values, which is
$\sup_{x\in \mathbb{R}}\left|l_\delta(x)-x\right|\leq C_0\delta$
for some positive constant $C_0$. Together with the sharpness of $F(x)$ given in \eqref{sharpness_F_pre}, we can conclude that, for some positive constant $C_1$,
$$\Delta(\bar{x})\leq C_1\delta,\ \forall \bar{x}\in \argmin_{x\in \mathbb{R}^p}F_\delta(x).$$
This inequality shows the closeness of minimums for $F(x)$ and $F_\delta(x)$. Moreover, our local landscape analysis, which aims at identifying and categorizing stationary points near the true signal vectors, allows for a finer characterization of local regions potentially containing a stationary point. When $p_{\text{fail}} = 0$, and under the conditions specified in \eqref{ineq_gen_sharp} and \eqref{rel_smooth_weak}, it is deduced that:
$$-\lambda_s\min\{\Delta(x),\frac{1}{\delta}\Delta^2(x)\}\geq F_\delta(x_\star)-F_\delta(x)\geq \nabla F_\delta(x)^\top (x_\star-x)-\frac{\rho}{2}\Delta^2(x).$$
So, when $\delta$ is sufficiently small, there exists positive constants $C_2,C_3$ such that
\begin{equation}\label{lower_bound_temp}
    \nabla F_\delta(x)^\top (x-x_\star)\geq C_2\min\{\Delta(x),\frac{1}{\delta}\Delta^2(x)\},
\end{equation}
$$\forall x\in \mathbb{R}^p\ s.t.\ \Delta(x)\leq C_3,\|x-x_\star\|_2 = \Delta(x).$$
Similar conclusions hold for the symmetrical area where $\Delta(z) = \|z+x_\star\|_2$. So, within $R_0 := \{z\in \mathbb{R}^p:\Delta(z)\leq C_3\}$, we can claim that true signal vectors are stationary points and local minima. 

In scenarios involving corruptions with $\gamma = 0$, we employ the decomposition
$$F_\delta(x) = (1-p_{\text{fail}})F_{\delta,1}(x) + p_{\text{fail}}F_{\delta,2}(x),$$
where $F_{\delta,i}(x)$, for $i=1,2$, represents the sample mean of $l_\delta\left((a_i^\top x)^2 - b_i\right)$ for the noiseless and corrupted measurements, respectively. This allows us to express the gradient as
$$\nabla F_\delta(x)^\top (x - x_\star) = (1-p_{\text{fail}})\nabla F_{\delta,1}(x)^\top (x-x_\star) + p_{\text{fail}}\nabla F_{\delta,2}(x)^\top (x - x_\star).$$
The first term on the right-hand side can be bounded in a manner akin to that in \eqref{lower_bound_temp}. For the second term, we will show in Section \ref{sec:landscape_noise} that, with high probability, its absolute value can be uniformly bounded by $C_4p_{\text{fail}}\|x-x_\star\|_2$ within $R_0$, where $C_4$ is a positive constant independent of $\delta$ and $p_{\text{fail}}$. This serves as an indication of the impact corrupted measurements have on the gradient. When $p_{\text{fail}}$ is sufficiently small, these bounds suggest that the vicinity $R_1$ around the true signal vectors within $R_0$ is defined as
$$R_1 = \{x \in \mathbb{R}^p : \Delta(x) \leq r_0\delta p_{\text{fail}}/(1-p_{\text{fail}})\},$$
for a positive constant $r_0$, and no stationary points are found in $R_0 \setminus R_1$. Therefore, if an algorithm minimizing $F_\delta(x)$ converges to a local minimum $x_{0,\star}$ within $R_0$, it must be located in $R_1$. This illustrates the robustness of employing $F_\delta(x)$, as the algorithm converges to a point near the true signal vectors. In a noiseless context, this point coincides with one of the true signal vectors. In cases with corruption, and when $\delta p_{\text{fail}}/(1-p_{\text{fail}})$ is small, the convergence point is close to either $\{x_\star, -x_\star\}$, serving as an effective warm-start for the IPL algorithm that minimizes $F(x)$ with notable local efficiency, thereby exactly finding one of the true signal vectors.

Addressing robustness alone is not sufficient given the nonconvex nature of $F_\delta(x)$. We also need to demonstrate the absence of spurious local minima outside the region $R_0$ to ensure the efficacy of algorithms with random initialization. In Sections \ref{sec:landscape_noiseless} and \ref{sec:landscape_noise}, we conduct landscape analysis to identify $R_0$ and to illustrate the benign landscape characteristics beyond $R_0$.

\textbf{Remark}: Inner product quantities akin to $\nabla F_\delta(x)^\top (x - x_\star)$, which involve the interaction between the (sub)gradient and the deviation from $x$ to the true vector, are pivotal in the landscape analysis across various nonconvex formulations \citep{davis2020nonsmooth, mei2018landscape, sun2018geometric, chen2019gradient}. Specifically, \cite{mei2018landscape} leveraged such quantities in the global landscape analysis of single-index models with the essential assumption of a strictly monotone link function. Given that phase retrieval, with its squared link function not satisfying this monotonicity requirement, usage of inner product quantities in our paper, \cite{sun2018geometric} and \cite{chen2019gradient} is confined to local vicinities. In these works, nonlocal landscape analyses are conducted through the examination of Hessian matrix-related quantities. A distinctive aspect of our study is the incorporation of corruption into the analysis, setting it apart from existing literature. The comprehensive details of this approach will be elaborated upon in Section \ref{sec:landscape_noise}.

Finally, we provide the following normal assumption that is stronger than both \ref{ass_direct} and \ref{ass_subg}.
\begin{assumption}
\label{ass_gaussian}
For any $i\in [n]$, $a_i\sim N(0,I_p)$.
\end{assumption}
Assumption \ref{ass_gaussian} was used in \cite{duchi2019solving,davis2020nonsmooth,sun2018geometric,chen2019gradient} to study initialization and landscape analysis for phase retrieval problems.

\subsection{Algorithms and Rate of Convergence}\label{subsec:methodology:rate}
In this subsection, we discuss the algorithms for minimizing $F_\delta(x)$. 
We will show that under the noiseless situation, with high probability, gradient descent on $F_\delta(x)$ enjoys local linear convergence. Before we start, we first give a Lemma that finds a Lipschitz constant $\mu_{max}/\delta$ for $\nabla F_\delta(x)$ with high probability.
\begin{lemma}\label{lip_F_delta}
Under Assumptions \ref{ass_data}, \ref{ass_kernel}, and \ref{ass_gaussian}, when $p_{\text{fail}} = \gamma = 0$, $\|x_\star\|_2=1$, and $p\geq 3$, there exist positive constants $\mu_{max},c,c',C$ such that, if $n\geq cp\log p$, then with probability at least $1-C\exp(c'n/\log n)- C/n$, we have
\begin{equation}\label{eq_lip_F_delta}
    \sup_{x\in\mathbb{R}^p}\|\partial^2 F_\delta(x)/\partial x\partial x^\top \|_2\leq \mu_{max}/\delta.
\end{equation}
Here, $\mu_{max}$ does not depend on $\delta$ but $c,c',C$ depend on $\delta$.
\end{lemma}
Given the above lemma, the local linear convergence for gradient descent can be shown in Theorem \ref{conv_noiseless_local}.
\begin{theorem}\label{conv_noiseless_local}
Suppose that \eqref{ineq_gen_sharp}, \eqref{rel_smooth_weak} and \eqref{eq_lip_F_delta} hold with $\lambda_s,\rho,\mu_{max}>0$, $p_{\text{fail}} = \gamma = 0$ and $\{x_k\}_{k=0}^\infty\subseteq \mathbb{R}^p$ is generated by gradient descent algorithm that iterates as follows:
$$x_{k+1} = x_k - t\nabla F_\delta(x_k),k = 0,1,2\ldots$$
with $t = \delta/\mu_{max}$. If $\Delta(x_0)\leq \min\left\{1,\sqrt{\frac{2\lambda_s}{\mu_{max}}}\right\}\delta$ and $0<\delta\leq \lambda_s/\rho$, we have
\begin{equation}\label{eq_conv_rate_gd}
    F_\delta(x_{k+1}) - F_\delta(x_\star)\leq (1-\lambda_s/(8\mu_{max})) \left(F_\delta(x_{k}) - F_\delta(x_\star)\right),\forall k\in \mathbb{N}.
\end{equation}
\end{theorem}
{In \eqref{eq_conv_rate_gd}, if we let $k = 0$ and choose $\Delta(x^0)>0$, we have $\lambda_s/(8\mu_{max})\leq 1$. Thus, we can define $\kappa := (8\mu_{max})/\lambda_s\geq 1$ as the condition number.} It is worth pointing out that {$\kappa$} is not related to $\delta$ or $p$ so the convergence rate is statistically stable. 

Next, we comment on the technical details we use for reaching this result. Linear convergence usually appears in a strongly convex situation. When investigating the Hessian matrix $\frac{\partial^2 F_\delta(x)}{\partial x\partial x^\top}$, $\frac{\mu_{max}}{\delta}$ provides an upper bound for $\left\|\frac{\partial^2 F_\delta(x)}{\partial x\partial x^\top}\right\|_2$. However, it is hard to find a positive lower bound for the smallest eigenvalue for the Hessian matrix even within the local vicinity of the true signal vectors. So, we turn to the Polyak-Lojasiewicz condition introduced in \cite{karimi2016linear}, which proves linear convergence of gradient descent algorithm for minimizing a function $f(x)$ by assuming
$\frac{1}{2}\|\nabla f(x)\|_2^2\geq \mu\left(f(x) - \min_{x} f(x)\right)$ 
with $\mu>0$.
We can prove that when \eqref{ineq_gen_sharp}, \eqref{rel_smooth_weak} and \eqref{eq_lip_F_delta} hold with $\lambda_s,\rho,\mu_{max}>0$, for any $\delta>0$ that is small enough, 
$$\frac{1}{2}\|\nabla F_\delta(x)\|_2^2\geq \frac{\mu_{min}}{\delta}\left(F_\delta(x) - F_\delta(x_\star)\right),\forall x\in \{x\in \mathbb{R}^p:\Delta(x)\leq C\delta\}$$
in which $\mu_{min}$ and $C$ are positive constants that are not related to the selection of $\delta$. With these conditions, we can get the convergence rate.

Linear convergence guarantees that minimizing the smoothed objective function $F_\delta(x)$ is not only sufficient for phase retrieval in the noiseless scenario but also offers enhanced efficiency in comparison to PL \citep{duchi2019solving} and IPL \cite{zheng2023new}. This improvement stems from the methodologies adopted by \cite{duchi2019solving}, which solves the subproblem utilizing the POGS algorithm \cite{parikh2014block}, and IPL, which solves the same subproblem through accelerated proximal gradient descent tailored for the dual problem. Both of these algorithms are characterized by their sub-linear convergence rates. Furthermore, a local linear convergence rate has been established across various phase retrieval algorithms, including Wirtinger flow \citep{candes2015phase}, truncated Wirtinger flow \citep{chen2017solving}, truncated amplitude flow \citep{wang2017solving}, and subgradient methods \citep{davis2018subgradient,davis2020nonsmooth}. Smoothed robust phase retrieval thus joins this list of methodologies. {In addition, Theorem \ref{conv_noiseless_local} and \eqref{ineq_gen_sharp} indicate that, for any $\varepsilon>0$, the vanilla gradient descent on the smoothed robust objective function $F_\delta(x)$ can find an $x$ such that $\Delta(x)\leq \varepsilon$ with at most $O(\kappa \log \frac{1}{\varepsilon})$ iterations. This is better than $O(\kappa^2 \log \frac{1}{\varepsilon})$ for the subgradient algorithms \cite{davis2018subgradient,davis2020nonsmooth} using the robust $\ell_1-$loss.}
In our numerical experiments in Section \ref{sec:numerical}, we employ monotone accelerated gradient descent (as outlined in Algorithm 1 of \cite{Li2015APG}) combined with line search techniques to achieve better overall efficiency. These experiments demonstrate that smoothed robust phase retrieval exhibits superior numerical efficiency on noiseless datasets when compared to both PL and IPL.

\subsection{Overview of Benign Landscape}
In this subsection, we provide an overview of the benign landscape for $F_\delta(x)$. Its gradient and Hessian matrix help us find and classify the stationary points, and we will discuss different kinds of benign landscapes in this paper.
\begin{enumerate}
    \item [(a)] \textbf{Noiseless population landscape.} In this case, $p_{\text{fail}} = \gamma = 0$ and we focus on the stationary point for $\mathbb{E}\left[F_\delta(x)\right]$. Under some assumptions, we exactly locate all the stationary points and prove that only $\{x_\star,-x_\star\}$ are local minimums.
    \item [(b)] \textbf{Noiseless empirical landscape.} In this case, $p_{\text{fail}} = \gamma = 0$ and we care about the stationary point of $F_\delta(x)$. Under some assumptions, we show that with high probability, the stationary points should be close to those for noiseless population landscape, and only $\{x_\star,-x_\star\}$ are local minimums.

    \item [(c)] \textbf{Empirical landscape with infrequent corruptions only.} In this case, $p_{\text{fail}} > 0,\gamma = 0$ and we study the stationary point of $F_\delta(x)$. Under some assumptions, we show that the stationary points are close to those of the noiseless population landscape and all the local minimums are close to $\{x_\star,-x_\star\}$ with high probability. Specifically, all the local minimums lie in the set
    $\left\{x\in \mathbb{R}^p:\Delta(x)\leq r_0\delta p_{\text{fail}}/(1-p_{\text{fail}})\right\}$
    in which $\Delta(x) = \mbox{dist}(x,\{x_\star,-x_\star\})$ and $r_0$ is a positive constant.  Thus, if an optimization algorithm finds a local minimum of $F_\delta(x)$, its gap to $\{x_\star,-x_\star\}$ is small when both $\delta$ and $p_{\text{fail}}$ are small.

    \item [(d)] \textbf{Empirical landscape with infrequent corruptions and bounded noises.} When $p_{\text{fail}}>0,\gamma>0$, similar results can be obtained when $\delta,\gamma,p_{\text{fail}}$ are small enough and $\delta\geq \Omega(\sqrt{\gamma})$, with high probability, all the local minimums lie in the set
    $\{x\in \mathbb{R}^p:\Delta(x)\leq \Tilde{r}_0\max\{\sqrt{\delta\gamma},\delta p_{\text{fail}}/(1-p_{\text{fail}})\}\}$
    where $\Tilde{r}_0$ is a positive constant.
\end{enumerate}

Now, we describe the procedure of smoothed robust phase retrieval. We first apply some gradient-based algorithm to minimize $F_\delta(x)$ and find a local minimum $x_{0,\star}$. Due to the benign landscapes described above, for the optimization, we expect that both random initialization and modified spectral initialization work well such that $x_{0,\star}$ is close to $x_\star$ or $-x_\star$. Second, for the noiseless situation, we can claim that $x_{0,\star}$ is already a true signal vector. For the situation with noise, benign landscape guarantees that $x_{0,\star}$ is close to one of the true signal vectors and is treated as a warm-start for IPL that minimizes $F(x)$ for exact recovery.

\section{Landscape Analysis in the Noiseless Case}\label{sec:landscape_noiseless}
In this section, we focus on the noiseless case with $p_{\text{fail}} = \gamma = 0$ and analyze both population and empirical landscapes.  Without loss of generosity, we always assume that $\|x_\star\|_2 = 1$ throughout this section.
\subsection{Notations}
Denote
$L(x;a)=l_\delta((a^\top x)^2-(a^\top x_\star)^2),x,a\in \mathbb{R}^p.$ 
The gradient vector is $\frac{\partial L(x;a)}{\partial x}=\left\{2l'_\delta\left((a^\top x)^2-(a^\top x_\star)^2\right)aa^\top \right\}x
$ and the Hessian matrix
$\frac{\partial^2 L(x;a)}{\partial x\partial x^\top} =2l'_\delta((a^\top x)^2-(a^\top x_\star)^2)aa^\top +4(a^\top x)^2l''_\delta((a^\top x)^2-(a^\top x_\star)^2)aa^\top $.
The empirical and population versions of objective functions can be written as $F_\delta(x) = \mathbb{E}_n\left[L(x;a)\right]$ and $\mathbb{E}\left[F_\delta(x)\right]=\mathbb{E}\left[L(x;a)\right].$
They have continuous gradient vectors and Hessian matrices as 
$\nabla F_\delta(x) = \mathbb{E}_n\left[\nabla L(x;a)\right]$, $\nabla \mathbb{E}\left[F_\delta(x)\right] = \mathbb{E}\left[\nabla L(x;a)\right]$, $\frac{\partial^2}{\partial x\partial x^\top } F_\delta(x) = \mathbb{E}_n\left[\frac{\partial^2}{\partial x\partial x^\top } L(x;a)\right]$, and $\frac{\partial^2}{\partial x\partial x^\top} \mathbb{E}\left[F_\delta(x)\right] = \mathbb{E}\left[\frac{\partial^2}{\partial x\partial x^\top} L(x;a)\right].$
We will use the Hessian matrix to classify the stationary points and find the benign noiseless population landscapes for $\mathbb{E}\left[F_\delta(x)\right]$ and empirical landscapes for $F_\delta(x)$.

Next, we introduce three quantities that are used to characterize the landscape. Define
\begin{equation*}
    U_1(x;a)=x^\top\frac{\partial L(x,a)}{\partial x}=2(a^\top x)^2l'_\delta((a^\top x)^2-(a^\top x_\star)^2),
\end{equation*}
\begin{equation*}
    U_2(x;a)=x_\star^\top \left[\frac{\partial^2 L(x;a)}{\partial x\partial x^\top }\right]x_\star,
\end{equation*}
and
\begin{equation*}
    U_3(x;a)=x_\star^\top \frac{\partial L(x,a)}{\partial x}=2(a^\top x_\star)l'_\delta((a^\top x)^2-(a^\top x_\star)^2)(a^\top x).
\end{equation*}
Based on Assumption \ref{ass_gaussian}, it is easy to find that, for any given $x_0\in \mathbb{R}^p$ satisfying $\|x_0\|_2>0$, we have
$z^\top \frac{\partial \mathbb{E}\left[F_\delta\right]}{\partial x}(x_0) = 0$, $\forall\ z\in\{x\in \mathbb{R}^p:x^\top x_0 = x^\top x_\star = 0\}.$ {This implies that most directional derivatives in $\mathbb{R}^p$ will be 0 and we only need to consider the remaining two directions.} To be specific, 
$$\mathbb{E}\left[U_1(x;a)\right] = \mathbb{E}\left[U_3(x;a)\right] = 0\iff \frac{\partial \mathbb{E}\left[F_\delta(x)\right]}{\partial x} = 0.$$
For $U_2(x;a)$, it is easy to find that $\mathbb{E}\left[U_2(x_0,a)\right]<0$ is sufficient for concluding that the Hessian matrix of $\mathbb{E}\left[F_\delta(x)\right]$ at $x_0$ contains at least one negative eigenvalue, which indicates that $x_0$ is not a local minimum. Similar properties hold for their empirical counterparts:
$$\mathbb{E}_n\left[U_1(x;a)\right] \neq 0 \mbox{ or } \mathbb{E}_n\left[U_3(x;a)\right] \neq 0 \implies \frac{\partial \mathbb{E}_n\left[F_\delta\right]}{\partial x}(x) \neq 0.$$
$$\mathbb{E}_n\left[U_2(x;a)\right] < 0 \implies x\mbox{ is not a local minimum for } F_\delta(x).$$

\subsection{Population Landscape}
\begin{figure}
    \centering
    \includegraphics[width=90mm]{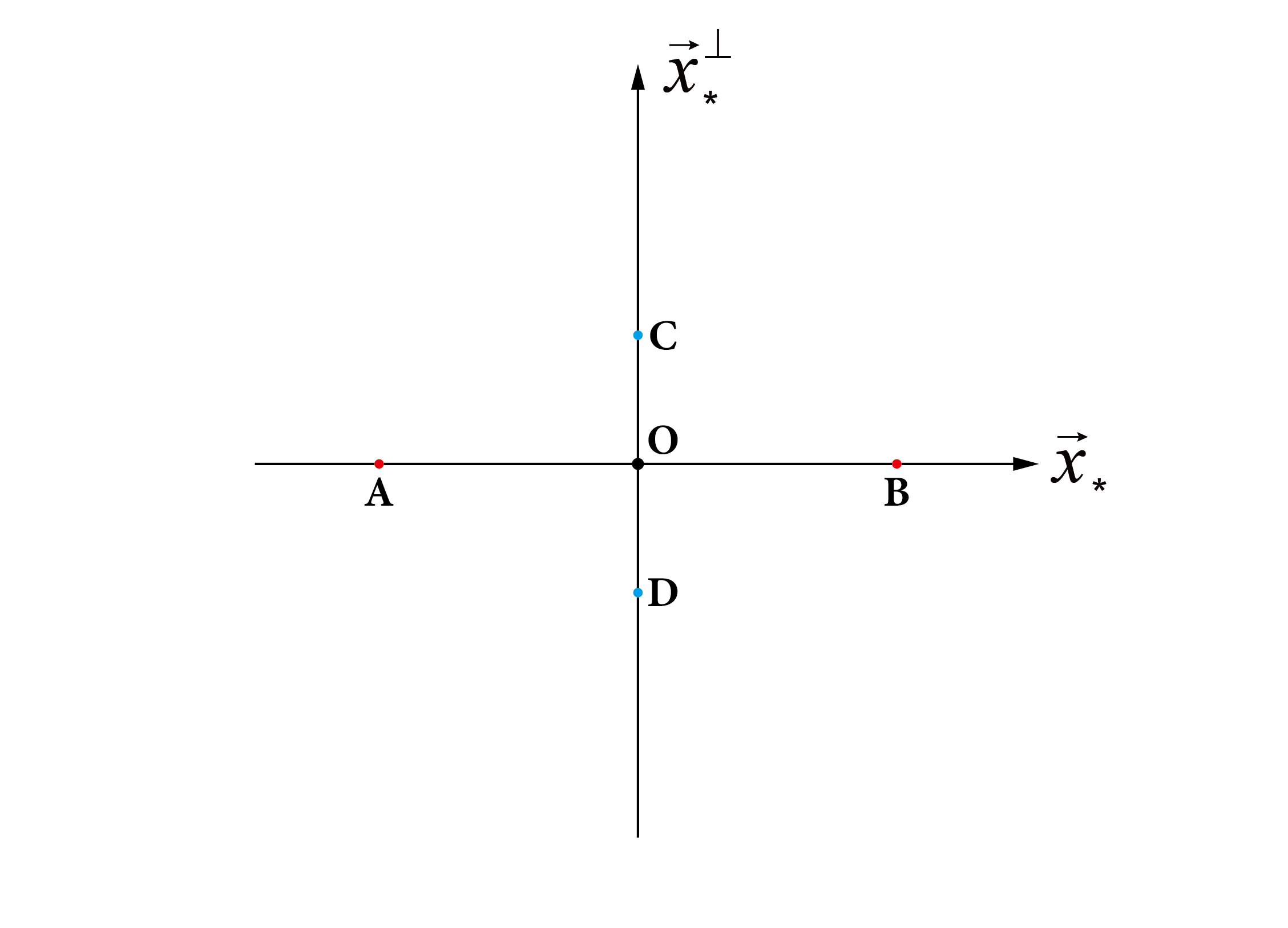}
    \caption{An illustration of the noiseless population landscape}
    \label{image_noiseless_pop}
\end{figure}

In this subsection, we will give the conclusions for the stationary points of $\mathbb{E}\left[F_\delta(x)\right]$. Denote $$S_0 = \{x\in \mathbb{R}^p: x^\top x_\star=0,\|x\|_2=u(\delta)\}$$
for some positive constant $u(\delta)$, whose rigorous definition will be given in Theorem \ref{theorem_position_stationary} later. To help to understand, we also use graphical explanations in Figure \ref{image_noiseless_pop}. We express vectors in the phase space $\mathbb{R}^p$ in a two-dimensional coordinate. For each vector $x\in \mathbb{R}^p$, we decompose it into the direction of $x_\star$ and an orthogonal direction as $x = ux_\star + vx_\star^\perp, x_\star^\top x_\star^\perp = 0,\|x_\star^\perp\|_2 = 1,$ with $u,v\in \mathbb{R}$. This decomposition exists for any $x\in \mathbb{R}^p$, and for a given $x$, $u$ and $|v|$ are unique. We let a point with coordinate $(u,v)$ denote the subset of $\mathbb{R}^p$ in which elements can give a decomposition with coefficients $u$ and $v$:
$$\left\{x\in \mathbb{R}^p:x^\top x_\star = u,\left\|x-(x^\top x_\star)x_\star\right\|_2 = |v|\right\}.$$
So, in Figure \ref{image_noiseless_pop}, if we let the first coordinate for $A,B$ be $-1$ and $1$ and the second coordinate for $C,D$ be $u(\delta)$ and $-u(\delta)$, we know that $A,B$ represent the true signal vectors, $C,D$ represent points in $S_0$ and the origin represents the $0\in \mathbb{R}^p$. For the noiseless population landscape, we can show that when $\delta< \delta_0$ with some positive constant $\delta_0$, there exists a unique positive constant $u(\delta)$ such that, the stationary point set is $S_0\bigcup \{0,x_\star,-x_\star\}$, among which, only global minimums $x_\star$ and $-x_\star$ are local minimums. This conclusion corresponds to Figure \ref{image_noiseless_pop}, in which $A,B,O,C,D$ are stationary points and only $A,B$ are local minimums. The following results describe the conclusion rigorously.
\begin{theorem}\label{theorem_position_stationary}
Under Assumptions \ref{ass_data}, \ref{ass_kernel} and \ref{ass_gaussian}, when $p_{\text{fail}} = \gamma = 0$ and $\|x_\star\|_2 = 1$, for any given $\delta>0$, there exists a unique $u(\delta)>0$ such that $\mathbf{S} := \left\{x\in\mathbb{R}^p:\frac{\partial \mathbb{E}\left[F_\delta(x)\right]}{\partial x}=0\right\}$, which is the stationary point set of $\mathbb{E}\left[F_\delta(x)\right]$, becomes
$$\mathbf{S} = \left\{0,x_\star,-x_\star\right\}\bigcup \left\{x\in\mathbb{R}^p:x^\top x_\star = 0,\|x\|_2 = u(\delta)\right\}.$$
\end{theorem}
The conclusion of Theorem \ref{theorem_position_stationary} is similar to that of \cite{davis2020nonsmooth}. Next, we comment on some key ideas behind the proof. It is easy to find that $\{0,x_\star,-x_\star\}$ are all stationary points. For $x\in \{x\in \mathbb{R}^p:x^\top x_\star = 0,\|x\|_2>0\},$ it is also easy to find out that $\mathbb{E}\left[U_3(x;a)\right]=0$. So we only need to solve the equation $\mathbb{E}\left[U_1(x;a)\right]=0.$ Based on the isotropy property of normal distribution, we can calculate $\mathbb{E}\left[U_1(x;a)\right]$ by assuming $x_\star = e_1$ and $x = ue_2,u>0$. So, $u(\delta)$ is the solution to the equation
$0 = \mathbb{E}\left[l'_\delta(u^2(a^\top e_2)^2-(a^\top e_1)^2)(a^\top e_2)^2\right]$ 
about $u>0$. For the complete proof, we also need to explain why $x$ is not a stationary point when $x^\top x_\star\neq 0$ and $\Delta(x) > 0$. The proof is a direct application of \cite{davis2020nonsmooth}'s work on spectral functions. Please refer to Appendix \ref{appendix:proof_positions}
for details.

With access to the Hessian matrix, we can also classify those stationary points. The conclusion is in the Theorem \ref{theorem_classify_stationary}.
\begin{theorem}\label{theorem_classify_stationary}
Under Assumptions \ref{ass_data}, \ref{ass_kernel}, \ref{ass_gaussian}, when $p_{\text{fail}} = \gamma = 0$ and $\|x_\star\|_2=1$, for $\mathbb{E}\left[F_\delta(x)\right]$, the following properties hold.
\begin{enumerate}
\item [(a)] $x_\star$ and $-x_\star$ are local and global minimums.
\item [(b)] $0\in \mathbb{R}^p$ is a local maximum.
\item [(c)] There exists a constant $\delta_0>0$ such that for any $\delta\in(0,\delta_0)$, 
$\mathbb{E}\left[U_2(x;a)\right]<0,\forall x\in S_0.$ This means that stationary points in $S_0$ are not local minimums.
\end{enumerate}
\end{theorem}
Theorem \ref{theorem_classify_stationary} classifies all the stationary points for $\mathbb{E}\left[F_\delta(x)\right]$ and indicates the benign noiseless population landscape that there is no spurious local minimum for $\mathbb{E}\left[F_\delta(x)\right]$. Another interesting result found in the proof of Theorem \ref{theorem_classify_stationary} is that the limits $\lim_{\delta\rightarrow 0}u(\delta)$ and $\lim_{\delta\rightarrow 0}\mathbb{E}\left[U_2(x;a)\right]$ do not depend on the selection of kernel $K(x)$. It shows that,
$$\lim_{\delta\rightarrow 0} u(\delta) = u_0\approx 0.44,$$
when $u_0$ is the unique solution of function $\frac{\pi}{4}=\arctan u+\frac{u}{u^2+1}$ and
$$\lim_{\delta\rightarrow 0}\mathbb{E}\left[U_2(x;a)\right] = \frac{16u^3}{\pi(1+u^2)^2}+\frac{8}{\pi}\left(\arctan(u)-\frac{\pi}{4}-\frac{u}{u^2+1}\right)$$
for any $x$ that satisfies $x^\top x_\star = 0$ and $\|x\|_2=u>0$. When $u = u_0$, $\lim_{\delta\rightarrow 0}\mathbb{E}\left[U_2(x;a)\right]\approx-1.57$, which is negative. This implies that these quantities are properties for the $\ell_1-$loss. 

Finally, we give a lemma about the positiveness of $U_1(x;a)$ when $\|x\|_2$ is very large.
\begin{lemma}\label{thm_infinite_pos_large}
Under Assumptions \ref{ass_data}, \ref{ass_kernel} and \ref{ass_gaussian}, when $p_{\text{fail}} = \gamma = 0$, there exists a positive constant $M>0$ such that 
$$\inf_{x\in \mathbb{R}^p,\|x\|_2 \geq M}\ \mathbb{E}\left[U_1(x;a)\right]>0.$$
\end{lemma}
Lemma \ref{thm_infinite_pos_large} further strengthens the non-zero nature of the gradient when $\|x\|_2$ is large and will be useful in reaching conclusions about the empirical landscape.

\subsection{Empirical Landscape}
In this subsection, we first show the conclusions for the benign empirical landscape and then provide a sketch of the proof. 

\begin{figure}
    \centering
    \includegraphics[width=90mm]{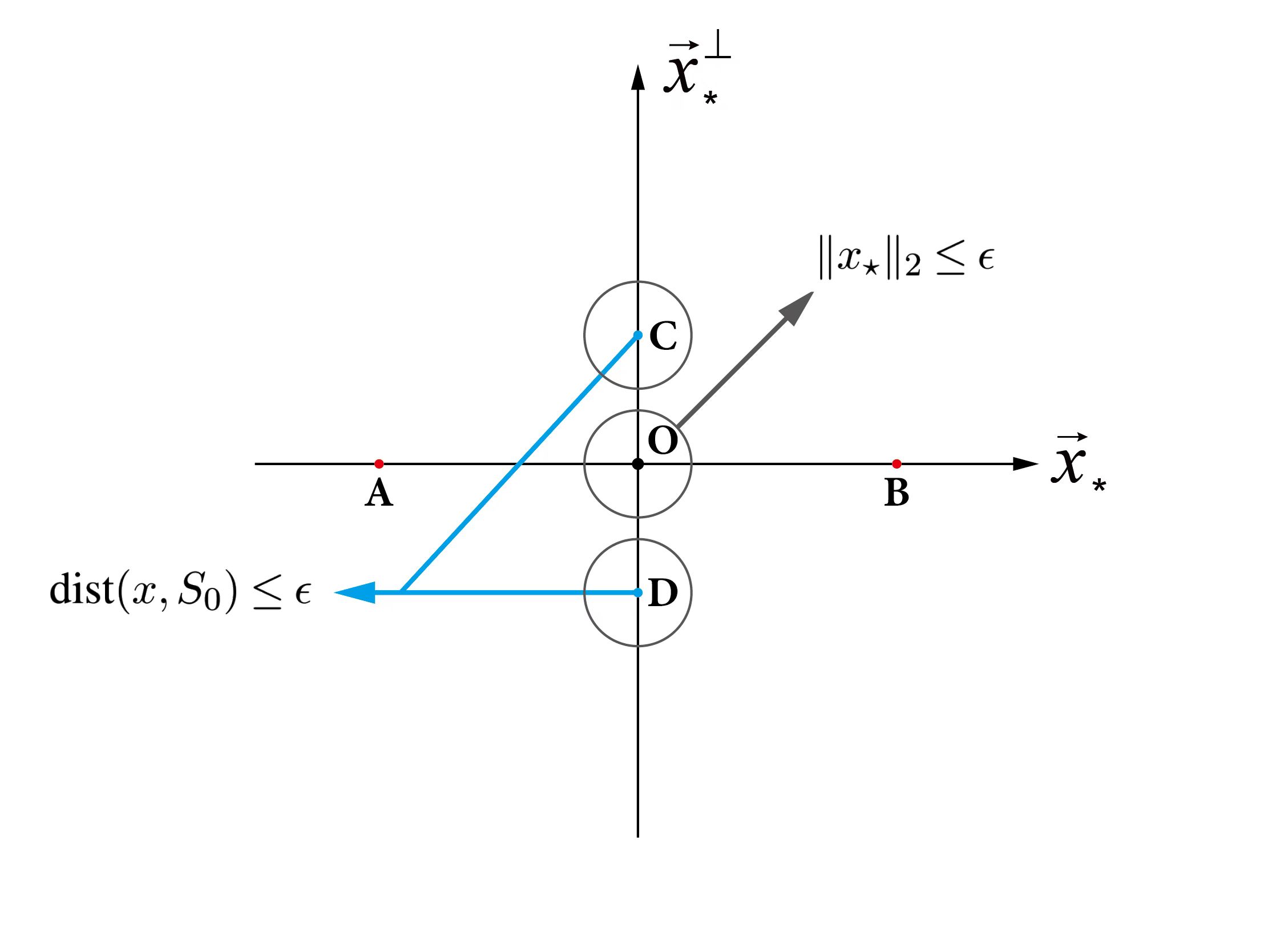}
    \caption{Noiseless Empirical Landscape}
    \label{image_noiseless_emp}
\end{figure}

We still use visualizations to help explain the landscape. Under some conditions, we can show that, with high probability, all the stationary points lie in the set
$$\mathbf{S}_1 = \{0,x_\star,-x_\star\}\bigcup \left\{x\in \mathbb{R}^p:\mbox{dist}(x,S_0\bigcup \{0\})\leq \epsilon\right\}$$
for any small positive constant $\epsilon$, among which, only global minimums $x_\star$ and $-x_\star$ are local minimums. In other words, a stationary point is either a true phase or lies in the vicinity of $S_0\bigcup\{0\}$. This conclusion corresponds to Figure \ref{image_noiseless_emp}, in which possible stationary points correspond to $A,B$ and the inner parts of the three circles, and only $A,B$ are local minimums.

Before giving the formal conclusion, we define the following events.
$$\mathcal{A}=\{ \textrm{there exists a stationary point of $F_\delta(x)$ such that it does not belong to the set $\mathbf{S}_1$} \}.$$
$$\mathcal{B}=\{ \textrm{there exists a local minimum of $F_\delta(x)$ such that it does not belong to the set $\{x_\star,-x_\star\}$} \}.$$
Formally, the conclusion is as follows.
\begin{theorem}\label{thm_land_noiseless}
Under Assumptions \ref{ass_data}, \ref{ass_kernel} and \ref{ass_gaussian}, when $p_{\text{fail}} = \gamma = 0$ and $\|x_\star\|_2=1$, there exists a constant $\delta_0>0$ such that, for any $\delta\in (0,\delta_0)$ and $\epsilon>0$, if $n\geq cp\log p$, we have
$$
P(\mathcal{A}\cup\mathcal{B})\le C\exp(-c'n)+\frac{C}{n},
$$
where $c,c',C$ are positive constants that depend on $\delta$ and $\epsilon$.
\end{theorem}

Theorem \ref{thm_land_noiseless} shows that the opposite event $\mathcal{A}^c\bigcap \mathcal{B}^c$ which corresponds to the benign noiseless empirical landscape happens with high probability. So, if we use a gradient-based optimization algorithm for minimizing $F_\delta(x)$ with random initialization and it finds a local minimum, the iterative sequence $\{x_k\}_{k=0}^\infty$ will converge to one of $\{x_\star,-x_\star\}$. In addition, under the benign landscape, though there are only theoretical guarantees for global convergence for first-order algorithms with additional tricks \citep{jin2017escape, jin2018accelerated} because of the existence of saddle points, numerical performances for robust phase retrieval indicate that we do not need those tricks for good performance. Readers with interests in saddle points can refer to \cite{jin2018accelerated} and references therein. The global convergence guarantee makes sure that our discussion of the local linear convergence rate in Section \ref{subsec:methodology:rate} is meaningful. 

For the rest of this subsection, we provide a sketch of the proof. In the first step, we focus on a large bounded area $R_2 = \{x\in \mathbb{R}^p:\|x\|_2\leq M\}$ and prove that all of $\mathbb{E}_n\left[U_i(x,a)\right],i = 1,2,3$ uniformly concentrate to their population versions with high probability. These quantities are related to the gradient and Hessian matrix for $F_\delta(x)$. If we do not consider the high-dimensional situation for $p$, it is easy to apply the central limit theorem to each element of the gradient and Hessian and claim convergence. However, if we consider the high-dimensional case when we can only assume that $n\geq \Omega(cp\log p)$, we can only focus on some selected key elements, which are $U_1(x;a),U_2(x;a)$ and $U_3(x;a)$. Fortunately, they are sufficient to support our claim about the empirical landscape. Next, we state our conclusions.
\begin{lemma}\label{thm_concen_noiseless}
Suppose that $\|x_\star\|_2 = 1$. Under Assumptions \ref{ass_data}, \ref{ass_kernel}, \ref{ass_gaussian}, for any given $\delta, M>0$ and $t>0$, when $n\geq cp\log p,p\geq 3$,$\|x_\star\|_2=1$, the following relationships hold.
\begin{equation}\label{eq_concen_u1_noiseless}
    P(\sup_{x\in B_p(0,M)}\left|\mathbb{E}_n\left[U_1(x;a)\right]-\mathbb{E}\left[U_1(x;a)\right]\right|\geq t)\leq C\exp\left(-c'n\right)+\frac{C}{n}.
\end{equation}
\begin{equation}\label{eq_concen_u2_noiseless}
    P(\sup_{x\in B_p(0,M)}\left|\mathbb{E}_n\left[U_2(x;a)\right]-\mathbb{E}\left[U_2(x;a)\right]\right|\geq t)\leq C\exp\left(-c'n\right)+\frac{C}{n}.
\end{equation}
\begin{equation}\label{eq_concen_u3_noiseless}
    P(\sup_{x\in B_p(0,M)}\left|\mathbb{E}_n\left[U_3(x;a)\right]-\mathbb{E}\left[U_3(x;a)\right]\right|\geq t)\leq C\exp\left(-c'n\right)+\frac{C}{n}.
\end{equation}
Here, $c\geq 1,c',C$ are positive constants that depend on $\delta,M,t$ but do not depend on $n$ or $p$. 
\end{lemma}
Lemma \ref{thm_concen_noiseless} guarantees the uniform concentration of the selected quantities within a large bounded subset of $\mathbb{R}^p$ when $n$ is large. This indicates that, within this area, stationary points and local minimums for $F_\delta(x)$ should be close to those for $\mathbb{E}\left[F_\delta(x)\right]$. It is also worth mentioning that, inferred from the proof, using smaller $\delta$ will require larger $n$ for these quantities to get close to their population version.

In the second step, we can go beyond closeness within a local area that contains the true signal vectors. Local landscape analysis strengthens the conclusion by showing that, within a local area that contains the true signal vectors, $\nabla F_\delta(x)$ is non-zero when $x\notin \{x_\star,-x_\star\}$, which means that there is no other stationary point in the local area. The two key properties for this part are the generalized sharpness condition given in Lemma \ref{lemma_sharpness_smooth} and the weak convexity condition given in Lemma \ref{lemma_smooth_weak_convexity}. With them, we can claim the following conclusion for the lower bound of the norm of the gradient and claim the benign local landscape.
\begin{lemma}\label{thm_noiseless_lower}
Suppose that $F_\delta(x)$ is weakly convex with parameter $\rho>0$ as given in Lemma \ref{lemma_smooth_weak_convexity} and also satisfies that
\begin{equation}\label{sharp_smooth_phase}
    F_\delta(x)-F_\delta(x_\star)\geq \lambda_s\min\{\Delta(x),\frac{1}{\delta}\Delta^2(x)\},\forall x\in \mathbb{R}^p
\end{equation}
as given in Lemma \ref{lemma_sharpness_smooth} for some positive constant $\lambda_s$. Then we have, for all $\delta\in (0,\lambda_s/\rho]$ and $x\in \{x\in \mathbb{R}^p:\|x-x_\star\|_2\leq \lambda_s/\rho\}$,
\begin{align}\label{smooth_sharp_low_u1u3}
    \mathbb{E}_n\left[U_1(x;a)\right]-\mathbb{E}_n\left[U_3(x;a)\right] &=\nabla F_\delta(x)^\top (x-x_\star)\geq \frac{\lambda_s}{2}\min\{\Delta(x),\frac{1}{\delta}\Delta^2(x)\},\nonumber
\end{align}
The symmetric case also holds when we replace $x_\star$ with $-x_\star$. After considering the symmetrical case based on $x + x_\star$, we have
$$\|\nabla F_\delta(x)\|_2>0,\ \forall\ x\in \{x\in \mathbb{R}^p:0<\Delta(x)\leq \lambda_s/\rho\}.$$
\end{lemma}

Finally, having handled $R_2$, we will prove that the unbounded area $\mathbb{R}^p\backslash R_2$ does not contain a stationary point by showing that $\mathbb{E}_n \left[U_1(x;a)\right]$ is positive on the boundary and that $\frac{1}{u^2}\mathbb{E}_n\left[U_1(ux_0,a)\right]$ is monotone increasing with regard to $u>0$ for any $x_0\in \partial R_2$. The monotonicity is formally given in Lemma \ref{lemma_monotone_u1}.
\begin{lemma}\label{lemma_monotone_u1}
For any constant $x_0\in \mathbb{R}^p,\|x_0\|_2>0,$ the function $h_{n}(u) = \frac{\mathbb{E}_{n}\left[U_1(ux_0;a)\right]}{u^2}$ is a monotone increasing function on $(0,+\infty)$.
\end{lemma}
Lemma \ref{lemma_monotone_u1} is obvious because $l'_\delta(\cdot)$ is monotone increasing. It directly indicates the following result about the unbounded area.
\begin{lemma}\label{thm_monotone_noiseless}
If there exists a constant $M>0$ such that
$\inf_{x\in \mathbb{R}^p:\|x\|_2=M}\mathbb{E}_{n}\left[U_1(x;a)\right]>0,$
then we have
$\inf_{x\in \mathbb{R}^p:\|x\|_2\geq M}\mathbb{E}_{n}\left[U_1(x;a)\right]>0.$
\end{lemma}
The positiveness of $\mathbb{E}_{n}\left[U_1(x;a)\right]$ means that there is no stationary point in this area.

\section{Empirical Landscape  in the Presence of Noise}\label{sec:landscape_noise}
In this section, we discuss the corrupted situation and the empirical landscape of $F_\delta(x)$. Without loss of generosity, we still assume that $\|x_\star\|_2 = 1$ throughout this section. We will follow the statistical assumptions in Section \ref{sec:methodology} with the general corruption setting $p_{\text{fail}} > 0$. Subsection \ref{subsec:noise:conclusions} will first provide the results under the situation that $\gamma = 0$ in Assumption \ref{ass_data} and then extend the result to the case that $\gamma\geq 0$. Subsection \ref{subsec:noise:sketch} will provide a sketch of the proof.
\subsection{Benign Landscape When Sparse Corruptions Exists}\label{subsec:noise:conclusions}
\begin{figure}
    \centering
    \includegraphics[width=90mm]{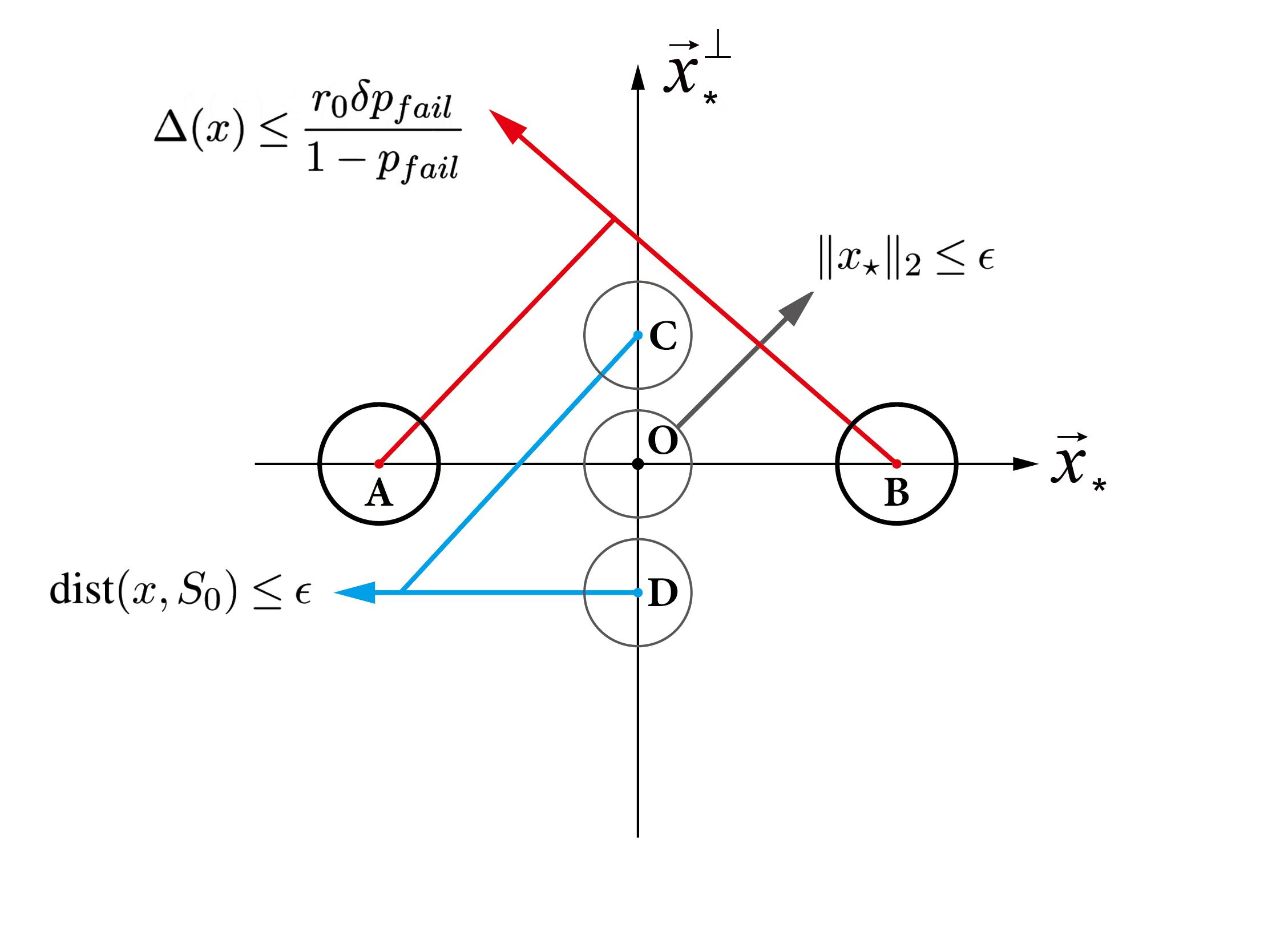}
    \caption{Empirical Landscape with Corruptions}
    \label{image_corrupted_emp}
\end{figure}

In this subsection, we first discuss the situation that $p_{\text{fail}}>0,\gamma = 0$ in Assumption \ref{ass_data}. We will show that, under some conditions, with high probability, all the stationary points of $F_\delta(x)$ lie in the set
$$\mathbf{S}_3 = \left\{x\in \mathbb{R}^p:\Delta(x)\leq r_0\delta p_{\text{fail}}/(1-p_{\text{fail}})\right\}\bigcup \left\{x\in \mathbb{R}^p:\mbox{dist}(x,S_0\bigcup \{0\})\leq \epsilon\right\}$$
for some positive constant $\epsilon$ that is related to $p_{\text{fail}}$ and a positive constant $r_0$, among which, local minimums lie in $$\mathbf{S}_4 = \left\{x\in \mathbb{R}^p:\Delta(x)\leq r_0\delta p_{\text{fail}}/(1-p_{\text{fail}})\right\}.$$ In other words, a stationary point either lies in the vicinity of $\{x_\star,-x_\star\}$ whose radius is proportional to $\delta p_{\text{fail}}/(1-p_{\text{fail}})$ or lies in the vicinity of $S_0\bigcup\{0\}$, and all local minimums lie in the vicinity of the true phase. This conclusion corresponds to Figure \ref{image_corrupted_emp}, in which possible stationary points correspond to the inner parts of the five circles and only inner parts of the circles $A$ and $B$ contain local minimums. The rigorous statement is given in the Theorem \ref{thm_land_corruption}.
\begin{theorem}\label{thm_land_corruption}
Suppose that $\|x_\star\|_2 = 1,p\geq 3$ and Assumptions  \ref{ass_data}, \ref{ass_kernel} and \ref{ass_gaussian} hold with $\gamma = 0$. We consider the following two events.
$$\mathcal{C}=\{ \textrm{there exists a stationary point of $F_\delta(x)$ such that it does not belong to the set $\mathbf{S}_3$} \}.$$
$$\mathcal{D}=\{ \textrm{there exists a local minimum of $F_\delta(x)$ such that it does not belong to the set $\mathbf{S}_4$} \}.$$
Then we have that for any $\epsilon>0$, there exist positive constants $\delta_0,r_0$ and $p_{\text{fail}}^m$, such that for any $\delta\in (0,\delta_0)$ and $p_{\text{fail}}\in (0,p_{\text{fail}}^m)$, when $n\geq cp\log p$, 
$$P(\mathcal{C}\bigcup \mathcal{D})\leq C\exp(-c'n/\log n)+\frac{C}{n}.$$ Here, the constants $c,c',C$  depend on $\delta, \epsilon$ and $p_{\text{fail}}$, and $r_0$ does not depend on $\delta, \epsilon$ or $p_{\text{fail}}$.
\end{theorem}
In Theorem \ref{thm_land_corruption}, we claim that the local minimums for $F_\delta(x)$ only exist in the vicinity of the true phase $\{x_\star,-x_\star\}$. The {radius $r_0\delta p_{\text{fail}}/(1-p_{\text{fail}})$ of the vicinity is proportional to the bandwidth and monotone increasing with regard to the proportion of the corrupted measurements.} In practice, we need to choose a $\delta$. Although smaller $\delta$ leads to closer local minimums, the non-local optimization efficiency is still unclear for difference choices of $\delta$, and decreasing $\delta$ will also require larger $n$ due to the dependency of the constants in Theorem \ref{thm_land_corruption} on $\delta$.
Next, we provide some guidance for picking $\delta$ under the situation that we do not know the covariance of $a_i$ and $\|x_\star\|_2$ is not necessarily 1. Using the consideration of homogeneity, we can let $\delta  = \delta_0\|\mathbb{E}_{n}\left[aa^\top \right]\|_2\|x_\star\|_2^2$, in which $\delta_0$ is a small but not extremely small positive constant. Finding or roughly estimating $\|\mathbb{E}_{n}\left[aa^\top \right]\|_2$ is easy in practice and a rough estimation of $\|x_\star\|_2$ can be done via prior knowledge or modified spectral initialization (Algorithm 3 in \cite{duchi2019solving}). We will provide an example of image recovery in Section \ref{sec:numerical}. For $\delta_0$ that should match the theoretical results based on the noiseless population landscape given in Theorem \ref{theorem_classify_stationary}, we will conduct sensitive analysis in Section \ref{sec:numerical}.

Next, we extend a benign landscape in  Theorem \ref{thm_land_corruption2} for the general situation that $\gamma\geq 0$.
\begin{theorem}\label{thm_land_corruption2}
Suppose that $\|x_\star\|_2 = 1,p\geq 3$. Under Assumptions \ref{ass_data}, \ref{ass_kernel} \ref{ass_gaussian}, there exist positive constants $\delta_0,\Tilde{r}_0$, $\gamma_0, c''$ and $p_{\text{fail}}^m$, such that when $n\geq cp\log p$, for any $\delta\in (0,\delta_0)$, $\gamma\in(0,\gamma_0)$ such that $\delta\geq c''\sqrt{\gamma}$ and $p_{\text{fail}}\in (0,p_{\text{fail}}^m)$, the probability that a local minimum of $F_\delta(x)$ exists and does not belongs to the set 
\begin{equation}\label{set_vicinity_local_gen}
    \{x\in \mathbb{R}^p:\Delta(x)\geq \Tilde{r}_0\max\{\sqrt{\delta\gamma},\delta p_{\text{fail}}/(1-p_{\text{fail}})\}\}
\end{equation}
 is at most $C\exp(-c'n/\log n)+\frac{C}{n}$. Here, $c,c',C$ are positive constants that depend on $\delta$ and $p_{\text{fail}}$, and $\Tilde{r}_0$ and $c''$ are positive constants that do not depend on $\delta$, $\gamma$ or $p_{\text{fail}}$.
\end{theorem}

Note that Theorem \ref{thm_land_corruption} is a special example of Theorem \ref{thm_land_corruption2} when $\gamma = 0$.

\subsection{Sketch of Proof}\label{subsec:noise:sketch}
We first provide a sketch of proof for Theorem \ref{thm_land_corruption}. Before we start, we define $   L(x;a,b)=l_\delta((a^\top x)^2-b),x\in \mathbb{R}^p$
and the definition of $F_\delta(x)$ in \eqref{loss_smoothed} can also be written as $F_\delta(x) = \mathbb{E}_{n}L(x;a,b).$ It is easy to find that $\frac{\partial L(x;a,b)}{\partial x}=\left\{2l'_\delta\left((a^\top x)^2-b\right)aa^\top \right\}x$,
and $\frac{\partial^2}{\partial x\partial x^\top } L(x;a,b)=2\left\{l'_\delta((a^\top x)^2-b)aa^\top \right\}+4\left\{(a^\top x)^2l''_\delta((a^\top x)^2-b)aa^\top \right\}.$ Similar to Section 3, we also define
\begin{equation*}
    U_1(x;a,b)=x^\top \frac{\partial L(x,a,b)}{\partial x}=2(a^\top x)^2l'_\delta((a^\top x)^2-b),
\end{equation*}
\begin{equation*}
    U_2(x;a,b)=x_\star^\top \left[\frac{\partial^2 L(x;a,b)}{\partial x\partial x^\top }\right]x_\star,
\end{equation*}
and
\begin{equation*}
    U_3(x;a,b)=x_\star^\top \frac{\partial L(x,a,b)}{\partial x}=2(a^\top x_\star)l'_\delta((a^\top x)^2-b)(a^\top x)
\end{equation*}

Now we provide the proof sketch. First, we focus on a large bounded area $R_2 = \{x\in \mathbb{R}^p:\|x\|_2\leq M\}$. The sample mean notations satisfies that $\mathbb{E}_n = (1-p_{\text{fail}})\mathbb{E}_{n_1} + p_{\text{fail}}\mathbb{E}_{n_2}$. {Here, $\mathbb{E}_{n_1}$ takes the sample mean with regard to indices from 1 to $n_1$, and $\mathbb{E}_{n_2}$ focuses on the indices from $n_1+1$ to $n_1+n_2$.} When $p_{\text{fail}}$ is small, We can expect that, in $R_2$, concentrations for the quantities defined above with $(1-p_{\text{fail}})\mathbb{E}_{n_1}$ hold as usual and those quantities with $p_{\text{fail}}\mathbb{E}_{n_2}$ are bounded by small positive constants. The rigorous statement is as follows.
\begin{lemma}\label{thm_concen_noise}
Under Assumptions \ref{ass_data}, \ref{ass_kernel} and \ref{ass_gaussian} with $\gamma = 0$, $p\geq 3,\|x_\star\|_2=1$, the following results hold.
\begin{enumerate}
    \item [(a)] For any given $M>0$, $t>0$ and $\delta>0$, when $n_1\geq c_1p\log p$, we have, $\forall i = 1,2,3,$
\begin{equation}\label{concen_corrupt_u1}
        P(\sup_{x\in B_p(0,M)}\left|\mathbb{E}_{n_1}\left[U_i(x;a,b)\right]-\mathbb{E}\left[U_i(x;a)\right]\right|\geq t)\leq C_1\exp\left(-c'_1n_1\right)+\frac{C_1}{n_1}.
    \end{equation}
Here, $c_1,c'_1,C_1$ are positive constants that depend on $\delta,M,t$.
\item [(b)] For any $M>0,t>0$ and $\delta>0$, when $n\geq c_2p\log p$, we have, $\forall i = 1,2,3$,
\begin{equation}\label{concen_corrupt_u1_upper}
    P(\sup_{x\in B_p(0,M)}\left|p_{\text{fail}}\mathbb{E}_{n_2}\left[U_i(x;a,b)\right]\right|\geq C'_2p_{\text{fail}}+t)\leq C_2\exp\left(-nc'_2/\log n\right)+\frac{C_2}{n}.
\end{equation}
Here, $c_2,c'_2,C_2,C'_2$ are all positive constants. $C'_2$ is affected by $M,\delta$ but is not affected by $t$ or $p_{\text{fail}}$. $c_2,c'_2,C_2$ are affected by $M,t,\delta$ but are not affected by $p_{\text{fail}}$.
\item [(c)] For any $M>0,t>0$ and $\delta>0$, when $n\geq c_3p\log p$, we have
\begin{equation}\label{concen_corrupt_u13_upper}
    P(\sup_{x\in B_p(0,M)}\left|p_{\text{fail}}\mathbb{E}_{n_2}\left[U_{1}(x;a,b) - U_{3}(x;a,b)\right]\right|\geq \left(C'_3p_{\text{fail}}+t\right)\|x-x_\star\|_2)
\end{equation}
$$\leq C_3\exp\left(-c'_3n\right)+\frac{C_3}{n}.$$
Here, $c_3,c'_3,C_3,C'_3$ are all positive constants. $C'_3$ is affected by $M$ but is not affected by $\delta$, $t$ or $p_{\text{fail}}$. $c_3,c'_3,C_3$ are affected by $M,t,\delta$ but are not affected by $p_{\text{fail}}$.
\end{enumerate}
 All the above conclusions also hold for the situation of replacing $x_\star$ with $-x_\star$.
\end{lemma}
Lemma \ref{thm_concen_noise}(a) is a direct conclusion from the noiseless case. For others, we can observe that, when $n$ is large, $t$ can be very small so that the corruptions affect each of $U_1(x;a,b),U_2(x;a,b)$ and $U_3(x;a,b)$ by some constant multiplying $p_{\text{fail}}$. This indicates that as long as $p_{\text{fail}}$ is smaller than some constant, the corruption will not overcome the power of normal measurements for $x$ that is not in the vicinity of $\{x_\star,-x_\star\}$. The last inequality will be used in the local landscape analysis to quantify the vicinity.

Second, we try to quantify the vicinity. Currently, within $R_2$, we know that all the local minimums are close to the true signal vectors. Local landscape analysis strengthens the conclusion by showing that, within a local area that contains the true signal vectors, $\nabla F_\delta(x)$ is non-zero when $x\notin \{x\in \mathbb{R}^p:\Delta(x)\leq r_0\delta p_{\text{fail}}/(1 - p_{\text{fail}})\}$, which quantifies the distance of the stationary points in the local area to the true signal vectors. The key of the analysis is the conclusion for the local landscape of the noiseless situation given in Lemma \ref{thm_noiseless_lower}, which means that with high probability, 
\begin{equation}\label{eq_norm_lower_noise}
\mathbb{E}_{n_1}\left[U_{1}(x;a,b)-U_{3}(x;a,b)\right]\geq \frac{\lambda_s}{2}\min\{\Delta(x),\frac{1}{\delta}\Delta^2(x)\},
\end{equation}
$$\forall \ x\in \{x\in \mathbb{R}^p:\|x-x_\star\|_2\leq \lambda_s/\rho\}.$$
Together with the third conclusion in Lemma \ref{thm_concen_noise}, we can find the following result for the local landscape. 
Lemma \ref{thm_local_land} quantifies the radius of the vicinity as $r_0\delta p_{\text{fail}}/(1-p_{\text{fail}})$. 
\begin{lemma}\label{thm_local_land}
When Assumptions \ref{ass_data}, \ref{ass_kernel} and \ref{ass_gaussian} hold, $\|x_\star\|_2=1,p\geq 3,\gamma = 0$, there exists positive constants $p_{\text{fail}}^m$ and $\delta_0$ such that when $n\geq cp\log p$, $0< p_{\text{fail}}\leq p_{\text{fail}}^m$ and $\delta<\delta_0$, the following relationship holds with probability at least $1-C\exp(c'n/\log n)-C/n$.
\begin{equation}\label{rel_local_land}
\mathbb{E}_{n}\left[U_{13}(x;a,b)\right]\geq C^{(1)}\min\{\Delta(x),\frac{1}{\delta}\Delta^2(x)\},
\end{equation}
$$\forall\ x\in \{x\in \mathbb{R}^p:r_0\delta p_{\text{fail}}/(1-p_{\text{fail}})\leq \|x-x_\star\|_2\leq C^{(2)}\}.$$
Here, $C^{(1)}$, $r_0$ and $C^{(2)}$ are positive constants that are not related to the selection of $\delta$ and $p_{\text{fail}}$. Positive constants $c,c',C$ are related to $\delta$ and $p_{\text{fail}}$. This conclusion also holds for the situation of replacing $x_\star$ with $-x_\star$.
\end{lemma}

Last, we will prove that the unbounded area $\mathbb{R}^p \backslash R_2$ does not contain a stationary point by showing that $\mathbb{E}_n U_1(x,a,b)$ is positive on the boundary and that $\frac{1}{u^2}\mathbb{E}_nU_1(ux_0,a,b)$ is monotone increasing with regard to $u>0$ for any $x_0\in \partial R_2$. The following two lemmas discuss monotonicity and positiveness similarly as in Lemmas \ref{lemma_monotone_u1} and \ref{thm_monotone_noiseless}.
\begin{lemma}
For any constant $x_0\in \mathbb{R}^p,\|x_0\|_2>0,$ the function $h_{n}(u) = \frac{\mathbb{E}_{n}\left[U_1(ux_0;a,b)\right]}{u^2}$ is a monotone increasing function on $(0,+\infty)$.
\end{lemma}
\begin{lemma}\label{thm_monotone}
If there exists a constant $M>0$ such that
$\inf_{x\in \mathbb{R}^p:\|x\|_2=M}\mathbb{E}_{n}\left[U_1(x,a,b)\right]>0,$
then we have
$\inf_{x\in \mathbb{R}^p:\|x\|_2\geq M}\mathbb{E}_{n}\left[U_1(x,a,b)\right]>0.$
\end{lemma}

{Finally, we discuss some key points for generalizing the proof of Theorem \ref{thm_land_corruption} for $\gamma = 0$ to Theorem \ref{thm_land_corruption2} for $\gamma>0$. First, requiring that $\mathbb{E}_{n_1}U_i(x;a,b)$ to be close to $\mathbb{E}_{n_1}U_i(x;a)$, $i=1,2,3$, we need that $\gamma$ is small and $\delta\geq \Omega(\sqrt{\gamma})$. Next, similar to the previous discussions on the sketch of proof, we need to let $(1-p_{\text{fail}})\mathbb{E}_{n_1}U_{13}(x,a,b)$ dominates $p_{\text{fail}}\mathbb{E}_{n_2}U_{13}(x,a,b)$ in some bounded area excluding the vicinity of $x_\star$. The estimation that $|p_{\text{fail}}\mathbb{E}_{n_2}U_{13}(x,a,b)|_2 = O(p_{\text{fail}}\|x-x_\star\|_2)$ with high probability remains unchanged. For the $(1-p_{\text{fail}})\mathbb{E}_{n_1}U_{13}(x,a,b)$, due to $\gamma>0$,  the relationship $\mathbb{E}_{n_1}U_{13}(x,a,b)\geq \Omega(\min\{\Delta(x),\Delta^2(x)/\delta\})$ only holds when $\|x-x_\star\|_2\geq \Omega(\sqrt{\gamma\delta})$. So, we can get the set given in \eqref{set_vicinity_local_gen} by comparing the bounds.}

\section{Numerical Experiments}\label{sec:numerical}
This section compares the numerical performance of PL\citep{duchi2019solving}, IPL\cite{zheng2023new}, and SRPR.  Subsection \ref{subsec:numerical:desc_algs} presents three algorithms. Subsections \ref{subsec:numerical:simulated} and \ref{subsec:numerical:real} present the numerical results from simulation and {a task of image recovery}, respectively.

\subsection{Description of Different Algorithms}\label{subsec:numerical:desc_algs}

To illustrate the efficacy of the benign landscape for smoothed robust phase retrieval, we evaluate both random initialization (RI) and modified spectral initialization (SI) as described in Algorithm 3 of \cite{duchi2019solving}. The SI method leverages spectral initialization on subsets of samples with small $b_i$ values, aiming to circumvent the influence of heavily-tailed corrupt measurements within $\{b_i, i = 1, 2, \ldots, n\}$. \cite{duchi2019solving} established that SI can, with high probability, generate an initial point $x_0$ satisfying $\Delta(x_0) \leq O(p_{\text{fail}})$, provided the ratio $n/p$ is sufficiently large and $p_{\text{fail}}$ is sufficiently small.

Next, we discuss the proximal linear (PL) algorithm presented in \cite{duchi2019solving} and the inexact PL (IPL) proposed in \cite{zheng2023new}. These algorithms are selected for comparative analysis due to PL's pioneering role in minimizing $F(x)$ and IPL's advancements, which enhance the numerical efficiency over PL and subgradient methods detailed in \cite{davis2018subgradient}. Thus they set a benchmark for robust phase retrieval \citep{zheng2023new}. Both PL and IPL proceed by solving a sequence of convex subproblems to minimize $F(x)$.

We then introduce the smoothed robust phase retrieval (SRPR) methodology. In the noiseless scenario, SRPR involves minimizing $F_\delta(x)$, utilizing either RI or SI for initialization. In cases involving corruption, SRPR entails an initial minimization of $F_\delta(x)$ using RI or SI, followed by employing the resultant minimization as a warm-start for IPL. Given that direct minimization of $F_\delta(x)$ does not yield an exact recovery of the true signal vectors, this two-step approach is necessary. For kernel selection across all comparisons with proximal linear algorithms, we employ the kernel $K(x) = \frac{1}{2(x^2 + 1)^{3/2}}$, which generates the Pseudo Huber loss function $l_\delta(x) = \sqrt{x^2 + \delta^2}$. The choice of $\delta$ will be elaborated upon subsequently. The monotone accelerated gradient descent algorithm (MAPG) with line search, as outlined in Algorithm 1 of \cite{Li2015APG}, is utilized for optimization. This algorithm, specifically designed for nonconvex optimization, incorporates a reference update mechanism based on direct gradient descent steps to ensure sufficient decrease. It effectively guarantees performance parity with, or superiority over, simple gradient descent. Subsequent numerical results will illustrate the linear convergence achievable through this algorithm in the noiseless context.

We consider the candidate algorithms in the comparison:
proximal linear algorithm with SI or RI (\textbf{PL-SI} and \textbf{PL-RI}), inexact proximal linear algorithm with SI or RI (\textbf{IPL-SI} and \textbf{IPL-RI}), and smoothed robust phase retrieval with SI or RI (\textbf{SRPR-SI} and \textbf{SRPR-RI}).

In what follows, we examine their performance on both simulated datasets and image recoveries. For an experiment with a specific dataset and initialization, let the final solution obtained by an algorithm be denoted as $x_K$. We consider the phase retrieval successful for simulated datasets if the relative error, defined as $\mbox{dist}(x_K, \{x_\star, -x_\star\}) / \|x_\star\|_2$, does not exceed $10^{-6}$ for both the IPL and SRPR algorithms. For PL, a less stringent threshold of $10^{-4}$ is adopted. For image recoveries, the success threshold for PL is adjusted to $10^{-1}$, whereas for IPL and SRPR, it remains unchanged. This differentiation in accuracy thresholds is attributed to the comparatively lower optimization efficiency of PL compared to IPL and SRPR. Next, we explain the CPU time associated with each algorithm. Specifically, it encompasses the aggregate duration spent by all components of an algorithm. For instance, the CPU time calculation for SRPR-SI in the context of corruptions includes the time required for modified spectral initialization, the minimization of $F_\delta(x)$ through MAPG, and the execution of IPL for exact recovery. These experiments are conducted on a server equipped with an Intel Xeon E5-2650v4 (2.2GHz) processor. To ensure robustness in our findings, each experimental setup, defined by a unique set of hyper-parameters related to the dataset, is replicated 50 times with varied random samples for all components involved. This procedure enables us to report both the success rate and median CPU time for each experimental scenario. Consistent with the observations made by \cite{duchi2019solving}, we note that an increase in $k = n/p$ and a decrease in $p_{\text{fail}}$ generally lead to higher success rates. However, the impact on CPU time does not lend itself to straightforward conclusions. The subsequent subsections will present numerical experiments that furnish further insights.

\subsection{Simulated Dataset}\label{subsec:numerical:simulated}
This subsection follows \cite{duchi2019solving} to conduct experiments on simulated datasets. We generate simulated datasets that $\{a_i,i=1,2\ldots n\}$ independently follows $N(0,I_p)$ and $x_\star\in\frac{1}{\sqrt{p}}\{-1,1\}^p$ with each element independently following discrete uniform distribution so that $\|x_\star\|_2=1$. The corruptions for these measurements are independent and the corruption type will be zeroing (i.e. letting $b_i=0,\forall i \in [n]\backslash [n_1]$) or using Cauchy distribution. A comprehensive description of the related hyper-parameters is as follows.
\begin{itemize}
\item Signal size for recovery: $p\in\{500,1500\}$.
\item Oversampling ratio $k=n/p\in\{2+0.25s|s\in[0,24],s\in N\}$.
\item Corruption type $r\in\{0,1\}$. $b_j = (a_j^\top x_\star)^2$ for $j\in [n_1]$. The corrupted response satisfies that for $\forall\ j\in [n]-[n_1]$, we have $b_j=r\tan (\frac{\pi}{2}U_j)\times \mbox{median}\left(\{(a_i^\top x_{\star})^2:i\in\{1,2,...,n\}\}\right),U_j\sim U(0,1)$ and are independent for different measurements. 
When $r=0$, the corrupted values are all equal to zero. This condition diminishes the effectiveness of the modified spectral initialization, which relies on $b_i$s with small magnitudes. Conversely, when $r=1$, the corrupted values are drawn from a positive Cauchy distribution, characterized by its infinite expected value. The choice of this distribution stems from its heavy-tailed nature as the capacity to handle heavy-tail corruption represents one of the robust model's benefits.

\item $p_{\text{fail}}\in \{0.025s|s\in [0,12],s\in N\}$.
\item Random initialization $x_0 = \Tilde{x}\Tilde{r}$, in which $\Tilde{x}\sim U[\partial B_p(0,1)]$ and $\Tilde{r}\sim U[0,4]$ are independent random variables.
\item $\delta = 0.25$ for SRPR when we make comparisons with PL and IPL, while $\delta$ can be selected easily here because we already know $\|x_\star\|_2$. 
\end{itemize}

\begin{figure}
\centering
\includegraphics[width=120mm]{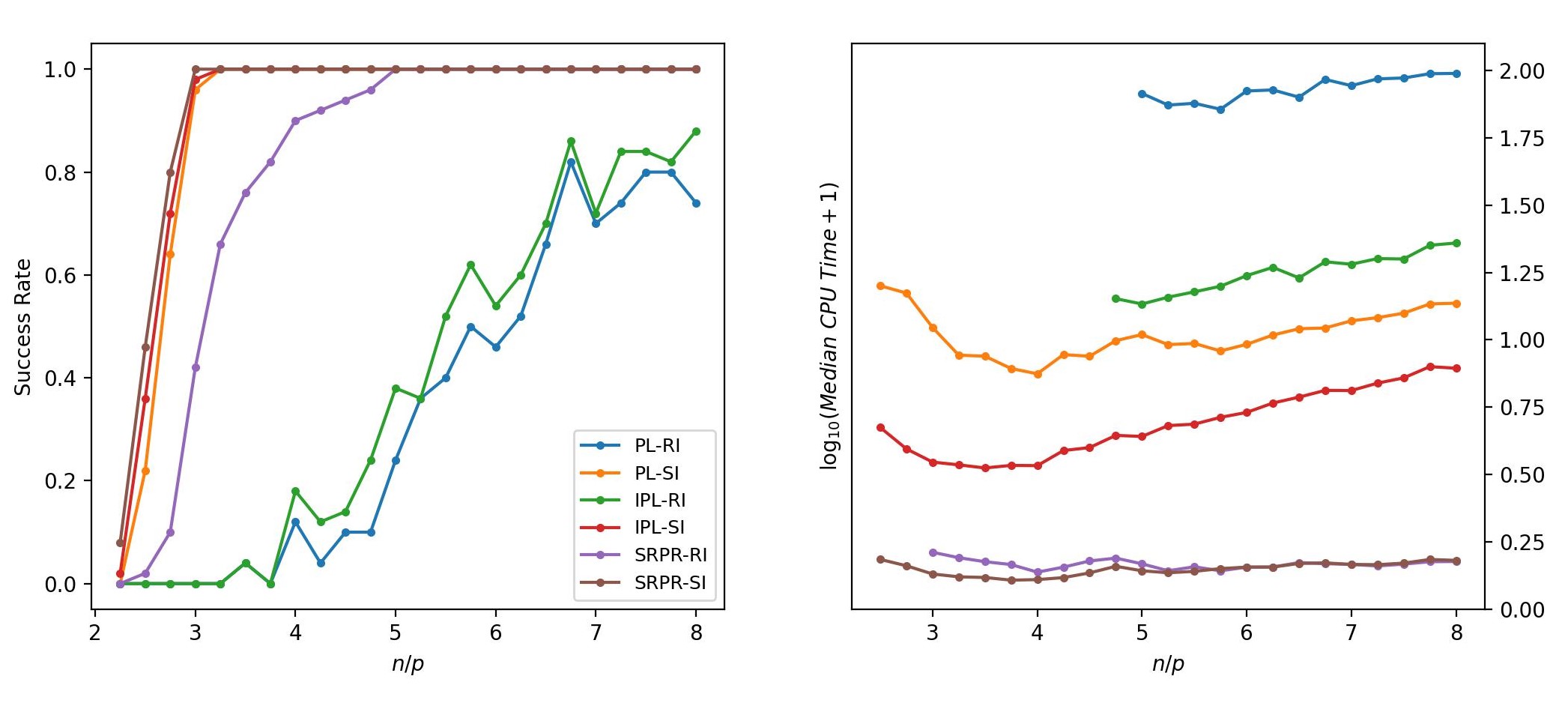}
\caption{Comparison of success rate and CPU time for the noiseless simulated datasets\label{generated_noiseless}}
\end{figure}

\begin{figure}
\centering
\includegraphics[width=120mm]{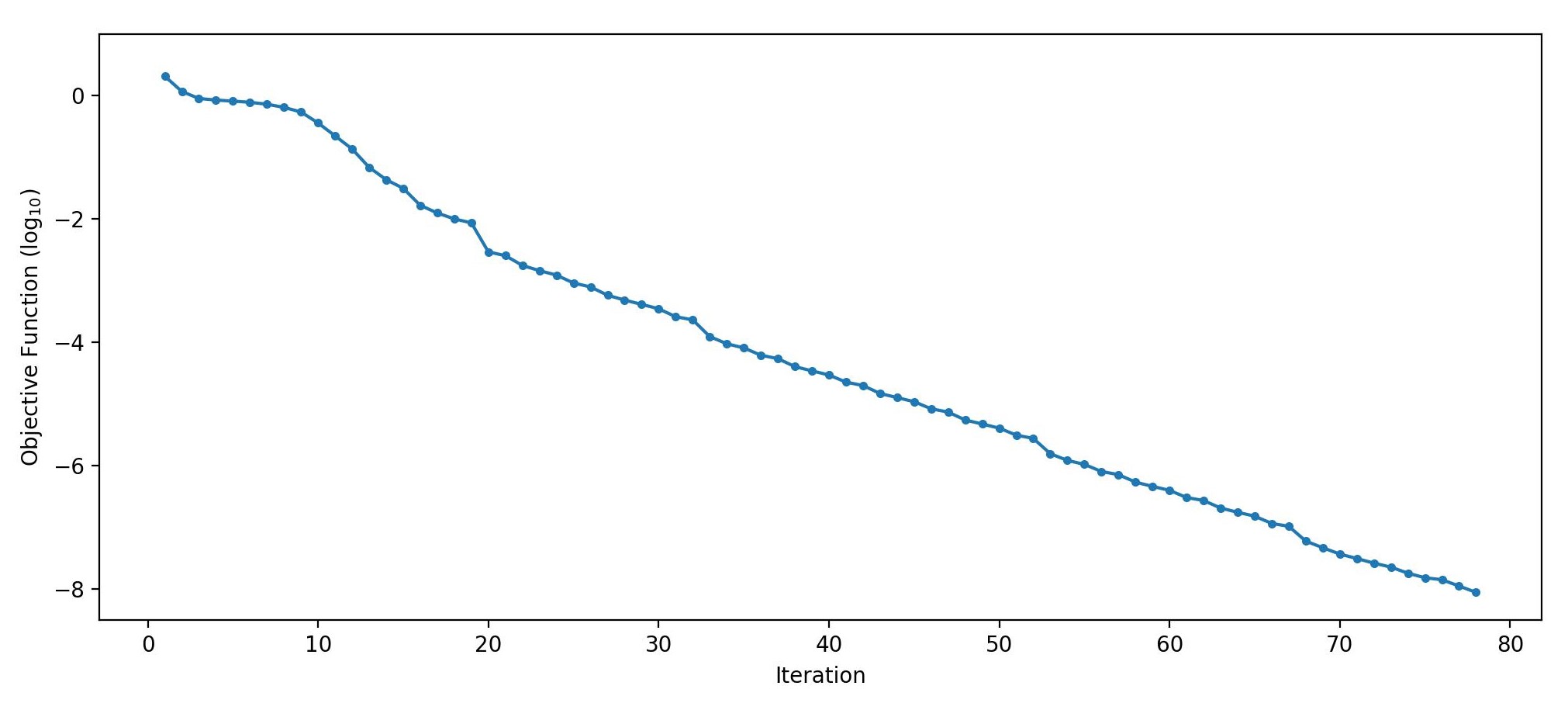}
\caption{Decreasing pattern of $F_\delta(x)$ with random initialization\label{obj_decrease} for the noiseless simulated dataset}
\end{figure}

Next, we present the performance of the candidate algorithms. Beginning with an evaluation under noiseless conditions, Figure \ref{generated_noiseless} illustrates the comparison between CPU time and success rate for these algorithms. The figure is divided into two panels: the left panel displays the success rate, while the right panel provides the median CPU time (after logarithmic transformation) for successful trials. An absence of data points indicates a success rate of zero. It is observed that algorithms incorporating SI exhibit superior success rates compared to those employing random initialization. Notably, SRPR-SI emerges as the most effective among them. For algorithms utilizing random initialization, SRPR achieves a $100\%$ success rate for large $n/p$ ratios, benefiting from the benign landscape. In contrast, PL and IPL fail to achieve similar success rates under these conditions. Regarding CPU time efficiency, SRPR stands out due to its local linear convergence property. This efficiency is further evidenced in Figure \ref{obj_decrease}, which delineates the decline of $F_\delta(x_k)$ over iterations for a trial with random initialization (where $p = 500$ and $k = 8$) using MAPG. The figure reveals a pronounced linear trend, with no discernible sub-linear or non-local stages.

Second, we focus on simulated datasets with corruptions. The conclusion of the comparison is consistent for all the combinations of hyper-parameters. We pick the combination $r=1, p=500,p_{\text{fail}} = 0.1$ and compare the performance of successful rate and CPU time for different selections of $n/p$.

\begin{figure}
\centering
\includegraphics[width=120mm]{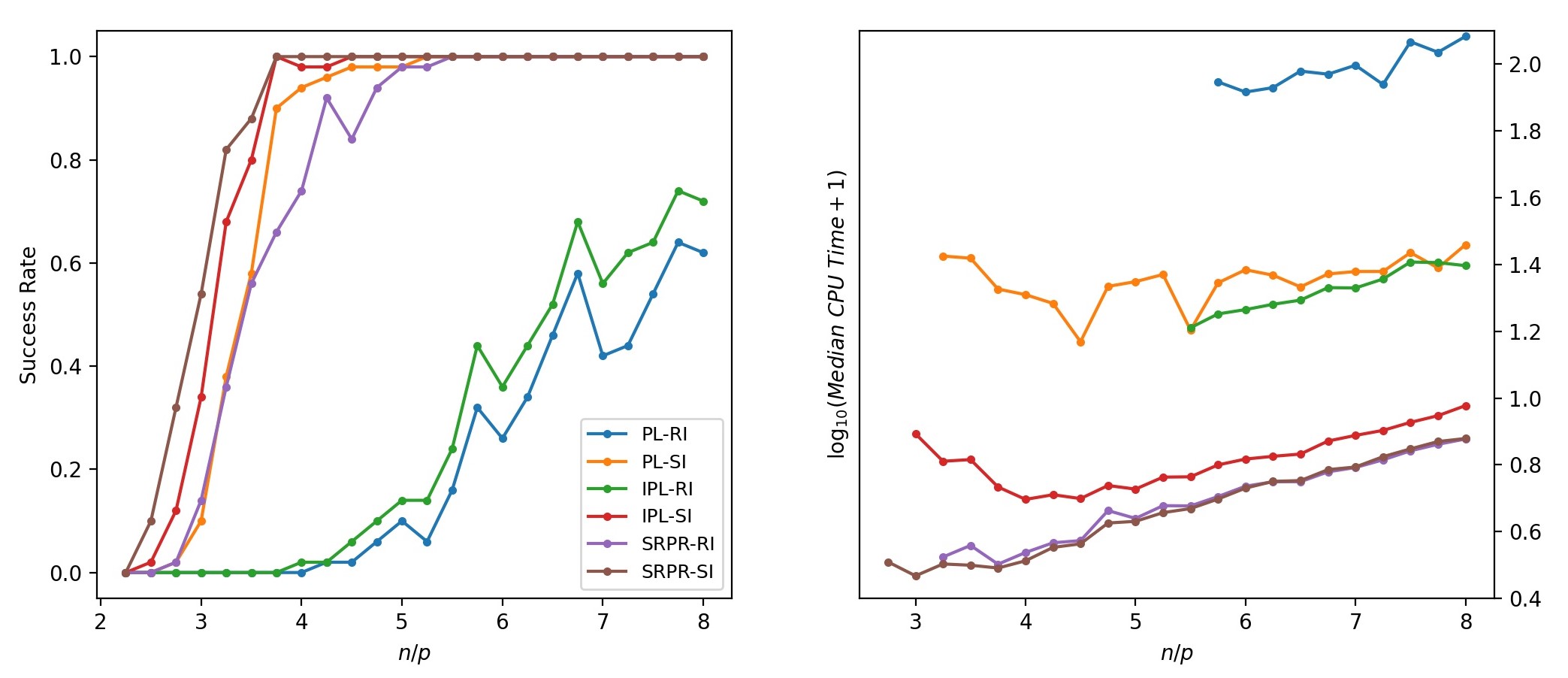}
\caption{Comparison of success rate and CPU time for corrupted simulated datasets\label{generated_corruption}}
\end{figure}

Figure \ref{generated_corruption} shows the performance of using these hyper-parameters and gives the same conclusion as Figure \ref{generated_noiseless} in terms of success rate. In terms of CPU time, there is only a little advantage for SRPR compared to IPL-SI as SRPR eventually turns back to minimizing $F(x)$ for exact recovery.

\begin{figure}
\centering
\includegraphics[width=120mm]{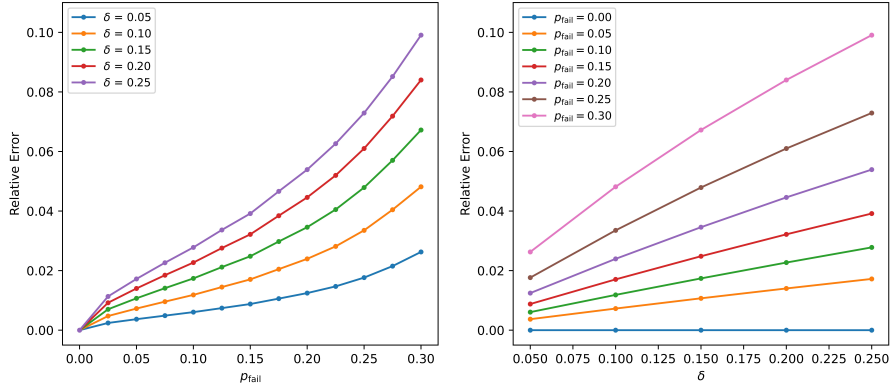}
\caption{Relative error after minimizing $F_\delta(x)$ for simulated datasets\label{relative_error}}
\end{figure}

In the sensitivity analysis concerning the parameter $\delta$, our focus centers on the relative error following the minimization of $F_\delta(x)$ across various settings of $\delta$ and $p_{\text{fail}}$. Employing hyperparameters $p = 1500$, $r = 1$, and $k = 8$, along with SI, Figure \ref{relative_error} delineates the outcomes from two perspectives. The first perspective plots $p_{\text{fail}}$ on the $x$-axis against the relative error on the $y$-axis, while the second uses $\delta$ as the $x$-axis. Each point on the graph represents the median relative error derived from 50 replicates. The analysis distinctly illustrates that lower values of $\delta$ and $p_{\text{fail}}$ are associated with reduced relative errors. This trend of increase in $\delta$ and $p_{\text{fail}}$ is in alignment with the theoretical results outlined in Theorem \ref{thm_land_corruption}. Notably, the precision achieved by $\min_{x\in\mathbb{R}^p } F_\delta(x)$ surpasses that of the SI, which exhibits a relative error ranging from 0.2 to 1.2.

\begin{figure}
\centering
\includegraphics[width=120mm]{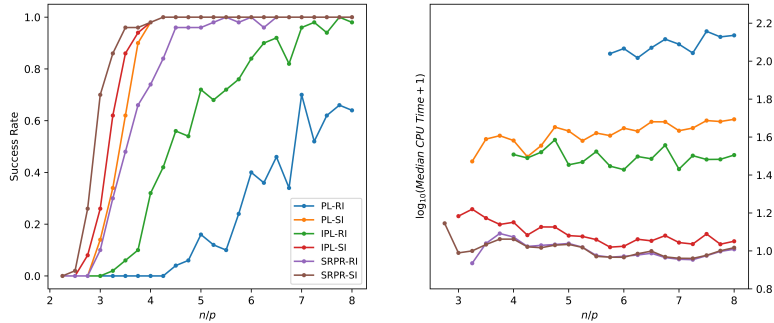}
\caption{Comparison for corrupted simulated datasets with bounded noise\label{generated_corruption2}}
\end{figure}

Next, we provide some numerical results that will show the robustness of the model when bounded noise exists. We let $b_j = (a_j^\top x_\star)^2+\tau_j, \forall j\in[n_1]$, in which $\tau_i$ is independent with $\{a_i\}_{i=1}^{n}$ and independently follows $U(-0.05,0.05)$. We set the threshold for success as 0.05. Figure \ref{generated_corruption2} shows the experiment results and shows similar patterns to Figure \ref{generated_corruption}. This means that SRPR still enjoys a benign landscape and advantages over PL and IPL. 

\subsection{Image Recovery}\label{subsec:numerical:real}
Now, we study the task of image recovery within the field of computational biology, leveraging the prevalent application of phase retrieval techniques \citep{huang2021holographic,stefik1978inferring}. Consider an image array $X_\star \in \mathbb{R}^{p_1 \times p_2 \times p_3}$, where each element takes a value within the interval $[0,1]$. Here, $p_1 \times p_2$ represents the image dimensions, and $p_3 = 3$ signifies the RGB color channels. We define the signal as $x_\star = [\mbox{vec}(X_{\star}); 0] \in \mathbb{R}^{p}$, where $p = \inf\{2^s : s \in \mathbb{N}, 2^s \geq p_1p_2p_3\}$, and $k = n/p \in \mathbb{N}$. The Hadamard matrix is denoted by $H_p \in \frac{1}{\sqrt{p}}\{-1,1\}^{p \times p}$, and we let $S_j \in \mbox{diag}(\{-1,1\}^p)$, for $j=1,2,\ldots,k$, be random diagonal matrices with diagonal elements following an independent discrete uniform distribution. The measurement matrix is defined as $A=[a_1^\top ; a_2^\top ; \ldots; a_{n}^\top ] = \frac{1}{\sqrt{k}}[H_pS_1; H_pS_2; \ldots; H_pS_k]$. Here, the semicolons mean that the matrix blocks are stacked vertically. This configuration benefits from an efficient computational complexity of $O(n\log p)$ for the transformation $Ay$ for $y\in\mathbb{R}^p$, paralleling the complexity of the discrete Fourier transform commonly encountered in phase retrieval scenarios with complex transformation matrices.
We use a real RNA nanoparticles image\footnote{https://visualsonline.cancer.gov/details.cfm?imageid=11167} which is also used in \cite{duchi2019solving} with an original image size of $2133\times 2133\times 3$.
We follow \cite{duchi2019solving}'s code\footnote{https://stanford.edu/~jduchi/projects/phase-retrieval-code.tgz} to conduct experiments on a colored sub-image and pick $p=2^{18}$ or $p=2^{22}$. A description of the related hyper-parameters is as follows. We consider the signal size for recovery $p\in\{2^{18},2^{22}\}$, oversampling ratio $k=n/p\in\{3,6\}$, corruption type $r\in\{0,1\}$, and $p_{\text{fail}}\in \{0.025s|s\in [0,8],s\in N\}$. The corrupted response satisfies that for $j\in[n]\backslash [n_1]$, we have $b_j=r\tan (\frac{\pi}{2}U_j)\times \mbox{median}\left(\{(a_i^\top x_{\star})^2:i\in\{1,2,...,n\}\}\right),U_j\sim U(0,1)$ and are independent for different measurements. The random initialization is $x=\frac{\sqrt{p}}{2}\Tilde{x}$, in which $\Tilde{x}\sim U[\partial B_p(0,1)]$. Next, we provide the choice of $\delta$. We can find that $\mathbb{E}_{n}\left[aa^\top \right] = \frac{1}{n}A^\top A = I_p/n.$ Based on the property of an RGB image, we assume that $\|x_\star\|_2 = c_0\sqrt{p},c_0\in (0,1)$ and $c_0$ is usually small when there is a large proportion of black pixels. So, $\delta$ can be selected as
$  \delta = \delta_0\left\|\frac{1}{n}A^\top A\right\|_2\|x_\star\|_2^2 = \frac{\delta_0c_0^2}{k}
$
for some small positive constant $\delta_0$. As a result, we choose $\delta = 0.01/6$, since  $k\leq 6$.

In our experiments on image recoveries, we observe consistent results across different settings of $r$. For illustration purposes, we select $r=0$. The success rates are reported in Table \ref{success_image18}, highlighting that SRPR uniquely does not depend on modified spectral initialization for large values of $n/p$. Regarding the median CPU time when $p_{\text{fail}} > 0$, the performances of SRPR-RI, SRPR-SI, and IPL-SI are comparable. The CPU time spans from 200 to 600 seconds for $p = 2^{18}$, and from 6000 to 15000 seconds for $p = 2^{22}$. The computational times for the other algorithms are significantly longer. Interestingly, in scenarios where $p_{\text{fail}} = 0$, IPL-SI demonstrates markedly improved efficiency over its performance in cases of corruption, surpassing both SRPR-RI and SRPR-SI in terms of CPU time efficiency.

\begin{table}[]
\centering
\begin{tabular}{|c|c|c|c|c|}
\hline
Success Rate & $p_{\text{fail}}\leq 0.15, k=3$ & $p_{\text{fail}}\leq 0.15, k=6$ & $p_{\text{fail}}> 0.15, k=3$ & $p_{\text{fail}}> 0.15, k=6$ \\ \hline
PL-RI        & 0                     & 0                     & 0                  & 0                  \\ \hline
IPL-RI       & 0                     & 0                     & 0                  & 0                  \\ \hline
SRPR-RI      & 0                     & 1                     & 0                  & 1                  \\ \hline
PL-SI        & 1                     & 1                     & 0                  & 1                  \\ \hline
IPL-SI       & 1                     & 1                     & 0                  & 1                  \\ \hline
SRPR-SI      & 1                     & 1                     & 0                  & 1                  \\ \hline
\end{tabular}
\caption{Comparison of Success Rate when $p=2^{18}$ for Image Recoveries\label{success_image18}}
\end{table}
\section{Conclusion}\label{conclusion}

In this paper, we aim to fill an important gap in the literature by analyzing the essential geometric structure of the nonconvex robust phase retrieval. We propose the SRPR based on a family of convolution-type smoothed loss functions as a promising alternative to the RPR based on the $\ell_1$-loss. We prove the benign geometric structure of the SRPR with high probability. Our result under the noiseless setting matches the best available result using the least-squares formulations, and our result under the setting with infrequent but arbitrary corruptions provides the first landscape analysis of phase retrieval with corruption in the literature. We also prove the local linear convergence rate of gradient descent for solving the SRPR under the noiseless setting. Numerical experiments on both simulation and a task of image recovery are presented to demonstrate the performance of the SRPR.

\bibliographystyle{agsm}
\bibliography{land_refs.bib}

\end{document}